\documentclass{article}
\usepackage{graphicx}

\PassOptionsToPackage{numbers, compress}{natbib}
 \usepackage[preprint]{neurips_2026}

\usepackage[utf8]{inputenc} 
\usepackage[T1]{fontenc}    
\usepackage{hyperref}       
\usepackage{url}            
\usepackage{booktabs}       
\usepackage{amsfonts}       
\usepackage{nicefrac}       
\usepackage{microtype}      
\usepackage{amsmath}
\usepackage{xcolor}         
\usepackage{algorithm}
\usepackage{algpseudocode}
\usepackage{setspace}
\usepackage{siunitx}

\usepackage{multirow}

\title{Mono-Forward: Revisiting Forward-Forward through Objective–Locality
Decomposition}

\author{%
  James Gong \\
  Department of Electrical, Computer, and Software Engineering\\
  University of Auckland\\
  Auckland, New Zealand \\
  \texttt{hgon777@aucklanduni.ac.nz}
  \And 
  Bruce Li \\
  Department of Electrical, Computer, and Software Engineering\\
  University of Auckland\\
  Auckland, New Zealand \\
  \texttt{tli389@aucklanduni.ac.nz}
  \And 
  Waleed Abdulla \\
  Department of Electrical, Computer, and Software Engineering\\
  University of Auckland\\
  Auckland, New Zealand \\
  \texttt{w.abdulla@auckland.ac.nz}
}

\begin{document}

\maketitle

\begin{abstract}
  Backpropagation remains the dominant algorithm for training deep neural networks, but it incurs substantial memory overhead and relies on global error propagation, which is often regarded as biologically implausible. The Forward-Forward (FF) algorithm is an appealing local-learning alternative to backpropagation, yet it still lags behind backpropagation in accuracy. A central unresolved question is whether this gap arises from FF’s locality or from the positive-negative double-pass goodness objective used to train each layer. In this work, we revisit FF under the supervised setting through a decomposition that separates these two design choices. Our analysis suggests that FF’s performance limitations are not explained by locality alone, but are also likely influenced by its goodness objective. Motivated by this view, we introduce Mono-Forward (MF), a simplification of FF that preserves its locality while replacing the contrastive goodness objective with a standard multi-class cross-entropy objective applied locally at each layer, serving as a controlled baseline for evaluating local learning under a standard classification objective. Across MLPs and convolutional networks, MF outperforms vanilla FF and remains competitive in multiple FF-variants. On MLP-Mixers, MF achieves stronger results on PathMNIST than backpropagation while requiring only 31\% of backpropagation's memory.
\end{abstract}

\section{Introduction}
\label{submission}
Backpropagation (BP) \cite{Rumelhart_Hinton_Williams_1986} is the dominant method for training deep neural networks \cite{LeCun_Bengio_Hinton_2015}. However, BP relies on the propagation of a global error signal throughout the entire network, a mechanism often considered biologically implausible and difficult to reconcile with local synaptic plasticity observed in the brain \cite{Song_Lukasiewicz_Xu_Bogacz_2020}.

Local learning algorithms are often motivated as alternatives to backpropagation because they avoid global error propagation and may offer advantages in memory usage, parallelization, and biological plausibility. Among local-learning methods, the Forward-Forward (FF) algorithm \cite{hinton2022forwardforwardalgorithmpreliminaryinvestigations} has attracted particular attention, and it replaces the conventional forward–backward training procedure with two forward passes and trains each layer using a local goodness-based objective. Despite this appealing structure, FF is substantially weaker than backpropagation in accuracy across many settings.

A key difficulty in interpreting this gap is that FF couples two distinct design choices: its locality and its contrastive goodness-based objective. As a result, when FF underperforms, it is unclear whether the limitation comes from locality itself or from the specific objective used to train each layer. This ambiguity makes it difficult to determine which aspects of FF should be preserved and which should be modified, when researchers work on refining it.

In this work, we revisit FF through a decomposition perspective. We separate FF into its local training structure and its learning objective, and use this separation to ask a central question: is FF weak because local learning is intrinsically limited, or because the original FF objective provides an inefficient training signal for classification tasks? Our ablations suggest the latter is a substantial part of the answer: when FF-style frameworks are combined with global error propagation, a notable gap to backpropagation remains, indicating that the objective itself is likely a bottleneck under the supervised setting.

To strengthen our evaluation, we propose Mono-Forward (MF), a simplification of FF rather than a completely new local-classifier paradigm. MF preserves FF’s layer-local learning structure but replaces the contrastive goodness objective with a standard multi-class cross-entropy objective. In this way, MF serves as a controlled test of the role played by the FF objective.

Our contributions are summarized as follows:
\begin{itemize}
    \item \textbf{Decomposition of the Forward-Forward Algorithm.}\\
    We provide a conceptual decomposition of FF into two components: its learning objective and its use of local error signals. Our analysis shows that the performance gap between FF and backpropagation is partly attributable to its learning objective rather than solely the locality of its updates.
    \item \textbf{Mono-Forward (MF): a minimal local-learning baseline.}\\
    We introduce MF, a simplified variant of FF that replaces the original contrastive objective with a standard cross-entropy classification objective applied locally at each layer. Importantly, MF serves as a controlled baseline that allows us to evaluate local learning under a standard classification objective, helping isolate the effect of the FF objective from the locality of its updates.
    \item \textbf{Improved accuracy with reduced memory.}\\
    Through experiments on MNIST, FashionMNIST, CIFAR-10, and CIFAR-100, we show that MF consistently improves over vanilla FF and remains competitive among multiple FF variants and other backpropagation-free algorithms. These results conceptually align with our decomposition analysis, suggesting that replacing the original FF objective with a standard classification objective can improve performance.
    \item \textbf{Broader evaluation across architectures and tasks.}\\
    We extend MF beyond standard MLP and CNN settings by implementing it on MLP-Mixer architectures. We further evaluate its performance across more challenging image classification tasks, including Tiny-ImageNet.
\end{itemize}

\section{Related Works}
The Forward–Forward algorithm, introduced by \citet{hinton2022forwardforwardalgorithmpreliminaryinvestigations}, offers a more biologically plausible alternative to backpropagation in multi‑layer perceptrons by replacing the single forward‑and‑backward pass with two distinct forward passes. Since its introduction, a number of works have attempted to improve the performance of the Forward-Forward algorithm by modifying its loss function, training dynamics, or architectural design. \citet{Lorberbom_Gat_Adi_Schwing_Hazan_2024} observe that the layer‑wise updates in the vanilla forward-forward algorithm can impede information flow between layers, and they therefore propose the collaborative forward-forward algorithm to enable richer inter-layer cooperation. The Symmetric Backpropagation-Free Contrastive (SymBa) Forward-Forward algorithm identifies inherent asymmetry in the Forward-Forward loss function and introduces an alternative loss function that results in faster convergence, while eliminating the threshold hyperparameter \cite{lee2023symbasymmetricbackpropagationfreecontrastive}. Meanwhile, a group of FF-based works reshape the algorithm into a self-supervised contrastive learning framework and learn a representation space that can be used for classification \cite{10491157, chen2025selfcontrastiveforwardforwardalgorithm, gong2025reshapingforwardforwardalgorithmsimilaritybased}. Unlike these approaches, which primarily propose alternative training mechanisms, our work focuses on analyzing the Forward-Forward algorithm itself by decomposing its learning objective and locality properties.

Other works on the Forward-Forward algorithm involve extending it beyond MLPs. These include the channel-wise forward-forward algorithm \cite{Papachristodoulou_2024}, adaptive spatial goodness encoding \cite{gong2026adaptivespatialgoodnessencoding}, distance-forward learning \cite{wu2024distanceforwardlearningenhancingforwardforward}, scalable forward-forward algorithm \cite{krutsylo2025scalableforwardforwardalgorithm}, DeeperForward \cite{sun2025deeperforward}, and the cascaded forward algorithm \cite{zhao2023cascadedforwardalgorithmneural}, all of which extend FF to convolutional neural networks (CNNs). The predictive forward-forward algorithm combines the predictive coding framework and applies the algorithm to recurrent neural networks (RNNs) \cite{ororbia2023predictiveforwardforwardalgorithm, predictive-coding2}. Our work extends FF to MLP-Mixers \cite{tolstikhin2021mlpmixerallmlparchitecturevision}, and, to the best of our knowledge, this is the first study to evaluate an FF-derived local-learning method within MLP-Mixer architectures.

The Forward-Forward algorithm can be viewed as a form of layer-wise learning, as it updates each layer of the network independently using locally computed objectives rather than propagating gradients through the entire network. More broadly, layer-wise learning methods aim to train deep networks by decomposing the global optimization problem into smaller local and layerwise learning tasks. Early work on layer-wise learning trained networks sequentially by optimizing one layer at a time until convergence before stacking the next \cite{NIPS2006_5da713a6}. This training was typically performed in an unsupervised manner to learn hierarchical feature representations, after which the entire network was fine-tuned using backpropagation for the target supervised task. A closely related paradigm is blockwise training, which partitions a deep network into several blocks and optimizes each block using local objectives. Unlike greedy layer-wise methods, blockwise approaches may train multiple layers jointly within each block and may allow different blocks to be optimized in parallel. Decoupled Greedy Learning (DGL) \cite{belilovsky2020decoupledgreedylearningcnns} is a representative example of this paradigm. DGL attaches auxiliary networks to intermediate blocks and optimizes them with local objectives while using a replay buffer to stabilize training, enabling blocks to be trained independently and asynchronously. Related blockwise training approaches have also been explored in the context of representation learning using self-supervised objectives \cite{siddiqui2024blockwiseselfsupervisedlearningscale, lwe2020puttingendendtoendgradientisolated}.

Beyond FF, a broad line of work seeks to reduce or eliminate backpropagation by replacing global gradient transport with alternative mechanisms. Representative examples include DTP \cite{lee2015differencetargetpropagation, ernoult2022scalingdifferencetargetpropagation}, which propagates target activations instead of gradients; FW-DTP \cite{shibuya2022fixedweightdifferencetargetpropagation}, which simplifies target propagation by using fixed random feedback weights; FA \cite{feedbackalignment}, which uses fixed random backward weights in place of exact transpose weights; DFA \cite{nøkland2016directfeedbackalignmentprovides}, which delivers fixed random output errors directly to hidden layers; CCL \cite{kao2024countercurrentlearningbiologicallyplausible}, which introduces a biologically plausible counter-current learning mechanism; SoftHebb \cite{Moraitis_2022}, which relies on local Hebbian-style synaptic updates; DRTP \cite{Frenkel_2021}, which projects supervisory signals to hidden layers through random mappings; PEPITA \cite{dellaferrera2023errordriveninputmodulationsolving}, which obtains local updates by perturbing the input with the output error; and rLRA \cite{10.1609/aaai.v37i8.26118}, which aligns intermediate representations to locally generated targets. While our work is not derived from these approaches, we discuss them to place our study of the Forward-Forward algorithm within the broader literature on alternatives to backpropagation.

\section{The Forward-Forward Algorithm}
In December 2022, \citet{hinton2022forwardforwardalgorithmpreliminaryinvestigations} proposed FF as an alternative to BP. FF, presented in Fig.~\ref{fig:forwardforwardvisual}, replaces the forward and backward passes in BP by two forward passes, termed the positive and negative forward passes. Each layer adjusts its parameters based on the "goodness" scores, evaluated separately in the positive and negative passes. In this work, we limit our study of FF to the supervised setting. A thorough discussion of FF under this setting is presented in Appendix~\ref{app:FF}.

In the rest of the paper, we use the term "learning objective" of FF frequently. Our definition of the learning objective of FF refers to the goal by which FF aims to produce high goodness values to positively encoded inputs during the positive pass and low goodness values to negatively encoded inputs during the negative pass. Under this definition, the learning objective includes the label-encoding procedure, the positive–negative dual-pass training setup, and the corresponding goodness-based loss function.

\section{Decomposing the Forward-Forward Algorithm}
\label{sec:decomposition}
The Forward-Forward algorithm combines two distinct design choices: (1) a contrastive goodness training objective, which consists of performing a positive forward pass and a negatively encoded forward pass and encouraging the goodness score of the positive pass to exceed that of the negative pass, and (2) layer-local learning, where each layer updates its parameters using only locally available information. In the original FF formulation, these two components are tightly coupled, making it unclear whether FF’s performance limitations arise from the locality of its updates or from the specific contrastive objective used to train each layer.

We further argue that the primary value of FF lies in its use of local error signals, which aligns with its intended motivation of biological plausibility. In contrast, its learning objective, particularly the reliance on negatively labeled data, raises concerns about biological plausibility, as there is little evidence that biological systems generate such explicitly negative samples \cite{negative-not1, negative-not2}.

By treating these components separately, we are able to investigate a central question: is the accuracy shortfall of FF really due to its use of local error signals, or is its learning objective also degrading performance? To explore this, we experiment with setups that retain the original FF learning objective but enable the propagation of a global error signal, training the network such that the final layer produces high goodness values for positive inputs and low goodness values for negative inputs. Our results in Table \ref{tab:ff_vs_ffge_bp} demonstrate that incorporating global error propagation into FF yields comparable accuracy on MNIST and FashionMNIST, while offering a modest improvement on CIFAR-10. However, all three datasets still exhibit a considerable gap compared to BP. This finding suggests that the current learning objective of FF inherently underperforms relative to the standard single-pass cross-entropy objective of BP. 

A likely explanation is that the FF objective is inherently a form of contrastive learning, and during a single update, FF only enforces that the goodness is higher for the correct label and lower for one particular incorrect label, but not necessarily for the other incorrect labels. The underlying hope is that, after iterating over the entire training set with randomly sampled negative encodings, the model will eventually learn to produce consistently higher goodness values for correct labels across the board. This limitation mirrors challenges seen in early contrastive frameworks, such as vanilla contrastive loss \cite{ccccccccc} and triplet loss \cite{Schroff_2015}, where only one negative example is considered per update. In those settings, accuracy and generalization can be significantly improved by incorporating a larger pool of negative examples or simply by using a classification loss that provides more structured supervision \cite{NIPS2016_6b180037, jacob2019metriclearninghordehighorder}. 

To further strengthen this ablation and enable a fairer comparison with BP, we enhance the FF learning objective so that the positive encoding is contrasted against all incorrect classes rather than a single sampled negative. Specifically, for a $C$-class problem, we include $C-1$ negative activations corresponding to every incorrect class during each update. This modification mirrors the multi-class cross-entropy objective used in backpropagation, which compares the correct class against all competing classes simultaneously.

Formally, let $f^+$ denote the goodness score associated with the correct class and $f_i$ denote the score for the $i$-th incorrect class. When all $C$-1 negatives are included, we define a full-comparison goodness objective as
\begin{equation}
    \mathcal{L}_\text{FC-FF} = \text{log}(1 + \sum^{C-1}_{i=1}\text{exp}(f_i - f^+)).
\end{equation}
This expression can be rearranged as
\begin{equation}
    \mathcal{L}_\text{FC-FF} = - \text{log}\frac{\text{exp}(f^+)}{\text{exp}(f^+) + \sum^{C-1}_{i=1}\text{exp}(f_i)}.
\end{equation}
The denominator corresponds to a partition function over all classes, which is identical to the normalization term in the softmax formulation. Therefore, when the negative set contains one example per incorrect class, this full-comparison goodness objective is algebraically equivalent to a softmax cross-entropy loss over the class goodness scores.
\begin{equation}
    \mathcal{L}_\text{CE} = -\text{log}\frac{\text{exp}(f^+)}{\sum^C_{j=1}\text{exp}(f_j)}.
\end{equation}

This shows that including all negative classes effectively transforms the FF objective into a softmax-like cross-entropy classification loss. We refer to this variant as Full-Comparison Forward-Forward (FC-FF). Similarly, we equip FC-FF with global error propagation (FC-FF + GE). The results in Table~\ref{tab:ff_vs_ffge_bp} show that accuracy gaps with BP still remain.

In the original FF algorithm, activations are \(L_2\)-normalized before being
passed to the next layer. This normalization was introduced to prevent later
layers from learning an identity-like mapping when an earlier layer already
assigns high goodness to positive samples and low goodness to negative samples.
By removing the activation magnitude, each subsequent layer is encouraged to
learn additional discriminative structure rather than simply propagating the
scale of the previous layer's goodness response.

However, in FF variants with global error propagation, such as FF + GE and
FC-FF + GE, the contrastive goodness objective is no longer enforced
independently at each layer. Instead, the objective is applied through the final
network output. Under this setting, the role of inter-layer \(L_2\)
normalization is less clear.

Let the pre-normalized activation at layer \(l\) be \(a_l\), and let the
activation passed to the next layer be
\[
    h_l = \frac{a_l}{\|a_l\|_2}.
\]
The mapping from \(a_l\) to \(h_l\) is invariant to positive rescaling, since
for any \(c>0\),
\[
    \frac{c a_l}{\|c a_l\|_2}
    =
    \frac{a_l}{\|a_l\|_2}.
\]
If the remainder of the network after layer \(l\) is represented by a function
\(F_l\), then the final prediction satisfies
\[
    \hat{y}
    =
    F_l(h_l)
    =
    F_l\left(\frac{a_l}{\|a_l\|_2}\right).
\]
Therefore,
\[
    \hat{y}(c a_l)
    =
    F_l\left(\frac{c a_l}{\|c a_l\|_2}\right)
    =
    F_l\left(\frac{a_l}{\|a_l\|_2}\right)
    =
    \hat{y}(a_l).
\]
Thus, inter-layer \(L_2\) normalization preserves only the direction of the
activation while discarding its radial component, \(\|a_l\|_2\).

From an information-theoretic perspective, a useful hidden representation should
preserve label-relevant information. Let \(X\), \(Y\), and \(H_l\) denote the
input, label, and hidden representation at layer \(l\), respectively. Since
\(H_l\) is computed from \(X\), the data-processing inequality gives
\[
    I(H_l;Y) \leq I(X;Y).
\]
Thus, a hidden representation cannot contain more label-relevant information
than is already present in the input. Directly estimating \(I(H_l;Y)\) for
high-dimensional neural representations is difficult. We therefore focus on a
narrower question: whether inter-layer \(L_2\) normalization removes information
that could be useful for predicting the label.

Although not all input-dependent information is label-relevant, the radial
component is discarded by construction rather than because it has been identified
as nuisance variation. If
\[
    I(\|a_l\|_2;Y \mid h_l) > 0,
\]
then the activation norm contains label-relevant information that is not
recoverable from the normalized direction alone. In this case, \(L_2\)
normalization restricts the information available to downstream layers. We
provide empirical evidence in Appendix~\ref{app:l2} that the radial component
contains nontrivial input-dependent variation.

Motivated by this observation, we introduce an additional ablation in which
inter-layer \(L_2\) normalization is removed while retaining both the
full-comparison objective and global error propagation. We denote this variant
as FC-NN-FF + GE. As shown in Table~\ref{tab:ff_vs_ffge_bp}, removing
normalization does not close the performance gap with backpropagation. In fact,
it deteriorates the performance of FC-FF + GE on both FashionMNIST and
CIFAR-10. These results suggest that inter-layer normalization is also not the 
cause of the performance gap.

To test this hypothesis more directly, we introduce MF, a simplified variant of
FF that removes the positive--negative multi-pass mechanism entirely. Although
the FC-FF objective can be algebraically related to a cross-entropy objective
when all negative classes are included, it still follows the original FF
training procedure, requiring separate positive and negative activations and
multiple forward passes. In contrast, MF replaces this mechanism with a standard
single-pass training procedure in which each layer is equipped with a local
classifier and optimized using a local multi-class cross-entropy objective. This
allows us to test whether replacing the FF objective and training procedure with
standard single-pass classification supervision improves learning while
retaining local parameter updates.

\begin{table*}[h!]
    \caption{Comparative accuracy (mean $\pm$ standard deviation) of FF, FF+GE (FF objective with global error), FC-FF + GE (Enhanced FF objective with global error), and BP on MNIST, FashionMNIST, and CIFAR-10 datasets using an MLP with 2 $\times$ 1000 hidden units. The networks were trained using an Adam optimizer, a fixed learning rate of $0.001$, and a batch size of 128. Each result is averaged over at least 3 random seeds. This experiment requires a peak memory of 200 MB and approximately 5 hours to finish with the same hardware setup described in Appendix~\ref{app:mlp-experiments}.}
    \vspace{7pt}
    \centering
    \begin{tabular}{c|c|c|c}
    \toprule
    Algorithm & MNIST (\%) & FashionMNIST (\%) & CIFAR-10 (\%) \\
    \midrule
    FF         & $97.35 \pm 0.07$ & $87.88 \pm 0.13$ & $48.59 \pm 0.28$ \\
    FF + GE    & $97.97 \pm 0.05$ & $87.64 \pm 0.08$ & $50.05 \pm 0.04$ \\
    FC-FF + GE & $98.10 \pm 0.08$ & $89.12 \pm 0.12$ & $53.27\pm 0.49$ \\
    FC-NN-FF + GE & $98.13 \pm 0.08$ & $88.23 \pm 0.22$  & $49.52 \pm 0.47$  \\ 
    BP         & $\textbf{98.69} \pm 0.05$ & $\textbf{90.42} \pm 0.19$ & $\textbf{58.41} \pm 0.26$ \\
    
    \bottomrule
    \end{tabular}
    \label{tab:ff_vs_ffge_bp}
\end{table*}

\section{The Mono-Forward Algorithm}
\label{sec:mono-forward}
Following the decomposition described in the previous section, we construct Mono-Forward as a minimal variant of the Forward-Forward algorithm in which the multi-pass contrastive goodness objective is replaced by a standard single-pass multi-class classification objective. 

\subsection{Training in MF}

MF, illustrated in Fig. \ref{fig:MF-training-figure}, extends the idea of goodness from the original FF algorithm in a simple and intuitive way. In FF, each layer produces a single goodness score that indicates how well the current representation matches a label-encoded input. In MF, this idea is generalized so that each layer produces a goodness score for every class. Therefore, instead of outputting only one scalar goodness value, each layer produces a \emph{goodness vector} whose entries indicate how strongly the current representation supports each possible class.

To achieve this, each layer $i$ is equipped with a layer-specific goodness matrix $M_i \in \mathbb{R}^{n \times m}$, where $n$ is the number of neurons in layer $i$ and $m$ is the number of classes. Given the activation vector $a_i$ of that layer, the class-specific goodness scores are computed as
\begin{equation}
    g_i = a_i^\top M_i,
    \label{eqn:MF-goodness}
\end{equation}
where $g_i \in \mathbb{R}^{m}$ is the goodness vector for a single example. Each element of $g_i$ represents the goodness of the current representation with respect to one class. For a batch of examples, these goodness vectors are stacked together to form the layer output matrix $G_i \in \mathbb{R}^{B \times m}$, where each row corresponds to one example and each column corresponds to one class. 

The layer output matrix is then used to compute a cross-entropy loss:
\begin{equation}
    \mathcal{L}_i = -\frac{1}{B}\sum^B_{b=1}\sum_{c=1}^m y_{bc} \log(\text{softmax}(G_{i})_{bc}),
\end{equation}
where $y_{bc}$ is a binary indicator showing whether example $b$ belongs to class $c$, and $B$ is the batch size. This loss encourages the goodness corresponding to the correct class to increase, while reducing the goodness assigned to incorrect classes. Intuitively, one can understand MF as appending a cross-entropy classifier to each network layer.

As illustrated in Fig. \ref{fig:MF-training-figure}, the loss computed from the layer output matrix at layer $i$ is used immediately to update the parameters of that layer, together with its layer-specific goodness matrix. After this local update, the detached layer activations $a_i$ are passed forward to the next layer.

\begin{table*}[t]
    \caption{Accuracy comparison on MLP benchmarks. Results are grouped by method family to distinguish FF-derived methods from non-FF local-learning baselines. The best result on each dataset is highlighted in bold. Vanilla FF fails to converge on CIFAR-100 (DNC). Experimental setup is documented in Appendix~\ref{app:mlp-experiments}. Reproduced results are indicated with *. NR stands for not reported.}
    \vspace{4pt}
    \centering
    \small
    \setlength{\tabcolsep}{5.5pt}
    \renewcommand{\arraystretch}{1.08}
    \begin{tabular}{lcccc}
    \toprule
    Method & MNIST (\%) & F-MNIST (\%) & CIFAR-10 (\%) & CIFAR-100 (\%) \\
    \midrule

    \multicolumn{5}{l}{\textbf{Backpropagation reference}} \\

    \midrule
    BP* 
    & $98.69 \pm 0.05$ 
    & $90.27 \pm 0.19$ 
    & $58.94 \pm 0.26$ 
    & $\mathbf{31.77} \pm 0.28$ \\

    \midrule
    \multicolumn{5}{l}{\textbf{FF-derived methods}} \\
    \midrule
    Vanilla FF* 
    & $97.35 \pm 0.07$ 
    & $87.88 \pm 0.13$ 
    & $48.75 \pm 0.12$ 
    & DNC \\

    FF + GE* 
    & $97.97 \pm 0.05$ 
    & $87.64 \pm 0.08$ 
    & $53.68 \pm 0.54$ 
    & $16.28 \pm 0.12$ \\

    FC-FF* 
    & $97.76 \pm 0.02$ 
    & $88.56 \pm 0.10$ 
    & $53.90 \pm 0.26$ 
    & NR \\

    FC-FF + GE* 
    & $98.10 \pm 0.08$ 
    & $89.12 \pm 0.12$ 
    & $54.02 \pm 0.28$ 
    & NR \\

    FFCM 
    & $97.12 \pm \mathrm{NR}$ 
    & $87.67 \pm \mathrm{NR}$ 
    & $54.48 \pm \mathrm{NR}$ 
    & NR \\

    SymBa 
    & $98.58 \pm \mathrm{NR}$ 
    & NR 
    & $\mathbf{59.09} \pm \mathrm{NR}$ 
    & $29.28 \pm \mathrm{NR}$ \\

    Collab FF 
    & $97.90 \pm 0.20$ 
    & $88.40 \pm 0.20$ 
    & $48.40 \pm 0.20$ 
    & NR \\

    FAUST 
    & $98.69 \pm \mathrm{NR}$ 
    & $89.67 \pm \mathrm{NR}$ 
    & $56.22 \pm \mathrm{NR}$ 
    & NR \\

    \midrule
    \multicolumn{5}{l}{\textbf{Non-FF baselines}} \\
    \midrule
    FA 
    & $98.36 \pm 0.03$ 
    & NR 
    & $53.60 \pm 0.40$ 
    & $26.20 \pm 0.30$ \\

    DFA 
    & $98.32 \pm 0.05$ 
    & NR 
    & $53.60 \pm 0.40$ 
    & $26.20 \pm 0.30$ \\

    DRTP 
    & $95.10 \pm 0.10$ 
    & NR 
    & $45.89 \pm 0.16$ 
    & $18.32 \pm 0.18$ \\

    FW-DTP 
    & $97.24 \pm 0.10$ 
    & $88.24 \pm 0.37$ 
    & $51.03 \pm 0.32$ 
    & $23.24 \pm 0.45$ \\

    CCL 
    & $98.13 \pm 0.10$ 
    & $88.58 \pm 0.29$ 
    & $52.73 \pm 0.59$ 
    & $21.76 \pm 0.22$ \\

    PEPITA 
    & $98.01 \pm 0.09$ 
    & NR 
    & $52.57 \pm 0.36$ 
    & $24.91 \pm 0.22$ \\

    \midrule
    \multicolumn{5}{l}{\textbf{Mono-Forward (Ours)}} \\
    \midrule
    MF cumulative* 
    & $\mathbf{98.74} \pm 0.01$ 
    & $\mathbf{90.52} \pm 0.02$ 
    & $58.24 \pm 0.17$ 
    & $31.36 \pm 0.13$ \\

    MF final* 
    & $\mathbf{98.74} \pm 0.02$ 
    & $90.51 \pm 0.05$ 
    & $58.30 \pm 0.20$ 
    & $31.50 \pm 0.25$ \\

    \bottomrule
    \end{tabular}
    \label{tab:accuracy_comparison_mlp}
\end{table*}

\begin{table*}[t]
    \caption{Accuracy comparison on CNN benchmarks. Results are grouped by method family to distinguish FF-derived methods from non-FF baselines. The best result on each dataset is highlighted in bold. Experimental setup is documented in Appendix~\ref{app:cnn-setup}. Reproduced results are indicated with *.}
    \vspace{4pt}
    \centering
    \small
    \setlength{\tabcolsep}{5.2pt}
    \renewcommand{\arraystretch}{1.08}
    \begin{tabular}{lccccc}
    \toprule
    Method & \# Layers & MNIST (\%) & F-MNIST (\%) & CIFAR-10 (\%) & CIFAR-100 (\%) \\
    \midrule

    \multicolumn{6}{l}{\textbf{Backpropagation reference}} \\
    \midrule
    BP* 
    & 4 
    & $99.52 \pm 0.05$ 
    & $93.63 \pm 0.06$ 
    & $86.22 \pm 0.49$ 
    & $\mathbf{70.37} \pm 0.16$ \\

    \midrule
    \multicolumn{6}{l}{\textbf{FF-derived methods}} \\
    \midrule

    Channel-wise FF 
    & 4 
    & $99.42 \pm 0.08$ 
    & $92.31 \pm 0.32$ 
    & $78.11 \pm 0.44$ 
    & $51.23 \pm \text{NR}$ \\

    CaFo 
    & 3 
    & $99.04 \pm \text{NR}$ 
    & NR 
    & $73.70 \pm \text{NR}$ 
    & $43.48\pm \text{NR}$ \\

    DeeperForward 
    & 4 
    & $99.50 \pm 0.05$ 
    & $91.83 \pm 0.06$ 
    & $79.49 \pm 0.29$ 
    & $43.48\pm \text{NR}$ \\

    FAUST 
    & 5 
    & $99.50\pm \text{NR}$ 
    & $93.12\pm \text{NR}$ 
    & $81.21\pm \text{NR}$ 
    & NR \\

    SCFF 
    & 5 
    & $98.70\pm \text{NR}$ 
    & NR 
    & $80.80\pm \text{NR}$ 
    & NR \\

    Scalable FF 
    & 3 
    & NR 
    & NR 
    & $81.38 \pm 0.35$ 
    & $55.34 \pm 0.07$ \\

    DF layerwise 
    & 10 
    & $99.53\pm \text{NR}$ 
    & $92.50\pm \text{NR}$ 
    & $84.75\pm \text{NR}$ 
    & $48.16\pm \text{NR}$ \\

    DF blockwise 
    & 10 
    & $\mathbf{99.70} \pm 0.09$ 
    & $93.89 \pm 0.25$ 
    & $88.15 \pm 0.28$ 
    & $59.01 \pm 0.35$ \\

    ASGE 
    & 11 
    & $99.59 \pm 0.06$ 
    & $92.98 \pm 0.24$ 
    & $\mathbf{90.62} \pm 0.17$ 
    & $65.42 \pm 0.16$ \\

    \midrule
    \multicolumn{6}{l}{\textbf{Non-FF baselines}} \\
    \midrule

    DFA* 
    & 4 
    & $98.53 \pm 0.17$ 
    & $87.87 \pm 0.21$ 
    & $57.53 \pm 0.19$ 
    & $24.54 \pm 0.60$ \\

    FA* 
    & 4 
    & $98.99 \pm 0.11$ 
    & $90.04 \pm 0.93$ 
    & $62.90 \pm 0.61$ 
    & $26.02 \pm 0.82$ \\

    PEPITA 
    & 2 
    & $98.29 \pm 0.13$ 
    & NR 
    & $56.33 \pm 1.35$ 
    & $27.56 \pm 0.60$ \\

    DRTP 
    & 2 
    & $97.32 \pm 0.25$ 
    & NR 
    & $50.53 \pm 0.81$ 
    & $20.14 \pm 0.68$ \\

    DTP 
    & 6 
    & $98.93 \pm 0.04$ 
    & $90.35 \pm 0.11$ 
    & $89.38 \pm 0.20$ 
    & NR \\

    CCL 
    & 5 
    & NR 
    & NR 
    & $82.94 \pm 0.53$ 
    & $56.29 \pm 0.25$ \\

    \midrule
    \multicolumn{6}{l}{\textbf{Mono-Forward (Ours)}} \\
    \midrule

    MF cumulative* 
    & 4 
    & $99.58 \pm 0.01$ 
    & $\mathbf{93.98} \pm 0.08$ 
    & $82.39 \pm 0.25$ 
    & $56.64 \pm 0.25$ \\

    MF final* 
    & 4 
    & $99.58 \pm 0.04$ 
    & $93.47 \pm 0.16$ 
    & $84.97 \pm 0.28$ 
    & $63.28 \pm 0.28$ \\

    \bottomrule
    \end{tabular}

    \label{tab:accuracy_comparison_cnn}
\end{table*}

\subsection{Prediction in MF}
At inference time, MF supports two prediction strategies. Given the goodness scores $\mathbf{G}_i$ at each layer $i$, the prediction can be obtained either by accumulating goodness scores across all layers or by using only the final-layer goodness:
\begin{equation}
\hat{y} =
\begin{cases}
\arg\max_c \sum_{i=1}^{L} g_{ic}, & \text{cumulative prediction},\\
\arg\max_c g_{Lc}, & \text{final-layer prediction}.
\end{cases}
\end{equation}
The final-layer strategy is more efficient because all intermediate goodness matrices used during training can be omitted at inference time. Consequently, inference uses a single forward pass through the backbone and only the final classifier head, matching the standard inference structure of a BP-trained network with a final linear head.

\subsection{Relationship to Prior Methods and Biological Learning Principles}
MF shares structural similarities with methods that attach auxiliary classifiers to intermediate layers or blocks, but its novelty is not the mere use of intermediate auxiliary goodness matrices; it is that MF is derived from a decomposition of FF, and used as a controlled test of the FF objective. In addition, MF is different from these methods in formulation and granularity. We discuss them in Appendix \ref{app:relationship-to-prior-methods}. A discussion of MF's biological plausibility is provided in Appendix \ref{app:biological-plausibility}.

\section{Experiments}
\label{experiments}
In this section, we compare the performance of MF against FF and standard BP across four benchmark datasets: MNIST, FashionMNIST, CIFAR-10, and CIFAR-100, summarized in Table~\ref{tab:accuracy_comparison_mlp}. All experiments in this section use the standard training and test splits provided with each dataset. Our results show that MF achieves accuracy close to BP, performing slightly better on MNIST and FashionMNIST, and slightly worse on CIFAR-10 and CIFAR-100, under the chosen MLP architectures. The results of FC-FF and FC-FF + GE on CIFAR-100 are omitted, due to the prohibitively high computational cost—requiring 99 negative passes and 1 positive pass per sample—which renders the approach impractical and of limited value for real-world applications.

To enable a broader comparison, we also evaluate MF against three recent extensions of the Forward-Forward algorithm on MLP: the Collaborative Forward-Forward algorithm (Collab FF) proposed by \citet{Lorberbom_Gat_Adi_Schwing_Hazan_2024}, FAUST \cite{gong2025reshapingforwardforwardalgorithmsimilaritybased}, and the Forward-Forward method with Contrastive Marginal loss (FFCM) proposed by \citet{10491157}. In addition, we include Feedback Alignment (FA) \cite{feedbackalignment}, Direct Feedback Alignment (DFA) \cite{nøkland2016directfeedbackalignmentprovides}, PEPITA \cite{dellaferrera2023errordriveninputmodulationsolving}, Counter-Current Learning (CCL) \cite{kao2024countercurrentlearningbiologicallyplausible}, Fixed-Weight Difference Target Propagation (FW-DTP) \cite{shibuya2022fixedweightdifferencetargetpropagation} and Direct Random Target Projection (DRTP) \cite{Frenkel_2021}, which all have implementations on fully connected MLPs, to provide additional comparison context. We report the best results as stated in the respective original papers, and to avoid claiming advantages from experimental settings, we refrain from using augmentation, dropout, or regularization.

To assess the performance of MF beyond fully connected architectures, we implement MF on a convolutional neural network and compare it against several Forward-Forward variants, including Channel-wise FF \cite{Papachristodoulou_2024}, DeeperForward \cite{sun2025deeperforward}, FAUST \cite{gong2025reshapingforwardforwardalgorithmsimilaritybased}, SCFF \cite{chen2025selfcontrastiveforwardforwardalgorithm}, and Scalable FF \cite{krutsylo2025scalableforwardforwardalgorithm}. We further include backpropagation-free alternatives such as FA, DFA, PEPITA, DRTP, CCL, and DTP \cite{ernoult2022scalingdifferencetargetpropagation}, together with standard backpropagation as a reference. DF \cite{wu2024distanceforwardlearningenhancingforwardforward} and ASGE \cite{gong2026adaptivespatialgoodnessencoding} are also closely related, as they are FF-variants, although their evaluations are conducted on deeper architectures. The reproduced results are evaluated using the same architecture, and these algorithms are selected because they have been previously evaluated on CNNs of similar depth. For methods where results are not reproduced, we report the best results from their original papers. The comparative results are summarized in Table~\ref{tab:accuracy_comparison_cnn}. Similar to the MLP experiments, we do not use augmentation, dropout, or regularization here.

Overall, our accuracy results indicate that MF is highly competitive in the pool of local learning algorithms, regardless of whether the algorithm is inspired by FF. This observation raises two important questions: do algorithms that underperform relative to MF in local learning settings also lag behind BP when global error propagation is enabled in the respective algorithms? If the global-error variants of these algorithms underperform, does it suggest that their underlying objectives may have room for further refinement\footnote{We recognize that only a subset of local-learning algorithms can be mechanistically equipped with global error; therefore this statement only applies to this subset of algorithms.}? We believe this question is a significant takeaway from this study; consequently, implementing and evaluating these global-error versions remains a key focus for our future work. In the following, we present more rigorous experimental analysis on MF against BP so that future studies on local learning can compare against it.

To assess the capability of MF beyond accuracy alone, we further examine the quality of the intermediate representations learned at each layer using three complementary metrics: linear-probe accuracy, k-NN accuracy, and Fisher ratio. We present our findings in Appendix \ref{app:representation-quality}. We also evaluate the convergence behavior of MF in Appendix~\ref{app:convergence-behavior}. To demonstrate that increasing the number of local iterations in MF is not beneficial, we conduct an ablation study on the effect of increasing local iteration $K$ in each layer. Our analysis is presented in Appendix~\ref{app:local-iter}.

One advantage of not using global error is its reduced memory usage during training. In layerwise operations, memory requirements depend primarily on the activations and computations of the current layer. To verify that MF retains this benefit under MLP settings, we analyze the memory usage in Appendix~\ref{app:memory-consumption}.


\subsection{Limitations of MF in CNNs and MF in MLP-Mixers}
\label{sec:cnn-limitation}
While MF can be applied to convolutional neural networks, doing so introduces an important computational challenge. We detail this limitation in Appendix~\ref{app:cnn-limitation}. Motivated by the computational limitations of MF on CNN architectures and its strong performance on MLP-based models, we further evaluate MF on MLP-Mixers, an architecture for computer vision composed entirely of MLP blocks \cite{tolstikhin2021mlpmixerallmlparchitecturevision}. We present our study on MLP-Mixers in Appendix~\ref{app:mlp-mixer}.

\section{Conclusions}
In this work, we revisited the Forward-Forward algorithm under the supervised setting through a decomposition of its two coupled ingredients: layer-local learning and the contrastive goodness objective. This perspective reframes a central question in local learning: whether FF underperforms because locality itself is limiting, or because FF couples locality with an inefficient objective. Our analysis suggests that the answer is not locality alone and the original FF objective is itself a substantial source of the performance gap to backpropagation. MF follows directly from this observation and serves as a principled simplification of FF that preserves its local learning structure while replacing the contrastive objective with a standard single-pass multi-class objective at each layer. In this sense, MF is both a practical training method and a controlled probe into what should be retained from FF and what should be redesigned. Across MLP benchmarks, MF consistently improves upon vanilla FF and recent FF variants, while achieving performance close to, and in some rare cases exceeding, backpropagation, with reduced memory cost.

In addition, our analysis reveals an important limitation in convolutional settings: because MF attaches class goodness matrices to intermediate representations, its memory cost can grow unfavorably with spatial resolution and the number of classes, exceeding backpropagation. As a natural progression, we implement MF on six different configurations of MLP-Mixers. Across all configurations, MF consistently achieves at least 58\% reduction in memory consumption compared to BP. While BP remains the stronger baseline method, we observe that MF slightly surpasses BP on PathMNIST while requiring only 31\% of the memory consumed by BP. This suggests that local learning objectives, when properly designed, can be advantageous in certain settings.

More broadly, this paper offers a simple perspective for analyzing local-learning methods. In particular, comparing a local-learning method with a variant that relaxes its locality constraint by introducing global error signals can help assess whether its limitations are more closely tied to locality or to the choice of training objective. While such comparisons may not always establish causality on their own, they can provide useful evidence for identifying the sources of performance differences in future studies of local learning.

\bibliographystyle{unsrtnat}
\bibliography{references}
\newpage
\appendix

\section{The Forward-Forward Algorithm}
\label{app:FF}
In FF, the positive goodness, evaluated during the positive pass, is computed as the sum of squared neuron activations for each layer $i$, with correctly labeled one-hot encoded data as input. Similarly, the negative goodness, obtained during the negative pass, is computed with incorrectly labeled one-hot encoded data as the input.\footnote{In the Forward-Forward framework, where there are \( m \) total classes, one-hot encoding is done by setting the first \( m \) pixels of the input image to black, except for the \( i_\text{th} \) pixel which is set to white. This modification encodes the label, signaling that the input corresponds to class \( i \).} The goal is to adjust the network parameters so that the positive goodness exceeds a specified threshold while the negative goodness remains well below it. Formally, the goodness values of a particular layer $i$ are
\begin{equation}
    G_{\text{pos,i}} = \sum_{n' = 0}^{n' = n - 1}(A_{i,n'}^2),\;\;\; x \rightarrow x_{\text{pos}}
    \label{eqn:positive_goodness}
\end{equation}
\begin{equation}
    G_{\text{neg,i}} = \sum_{n' = 0}^{n' = n - 1}(A_{i,n'}^2),\;\;\; x \rightarrow x_{\text{neg}}
    \label{eqn:negative_goodness}
\end{equation}
where $G_{\text{pos,i}}$ and $G_{\text{neg,i}}$ represent the layer-wise positive and negative goodness respectively. $A_{i,n'}$ represents the activation of the $n'_{\text{th}}$ neuron, where there are $n$ neurons in total. $x_{\text{pos}}$ and $x_\text{{neg}}$ indicate positive and negative input data.

The loss function used in the Forward-Forward framework is composed of two parts: a positive loss and a negative loss, each calculated during the positive pass and the negative pass, respectively. The positive and negative loss values for each layer \( i \) are computed as follows, with the threshold parameter \( \theta \):
\begin{equation}
    L_{\text{pos,i}} = \text{log}(1 + e^{\theta - G_{\text{pos,i}}}),\;\;L_{\text{neg,i}} = \text{log}(1 + e^{G_{\text{neg,i}} - \theta})
    \label{eqn:positive-negative-losses}
\end{equation}

The total loss for layer $i$ is then defined as the average of these two components:
\begin{equation}
    L_{\text{total,i}} = \frac{1}{2}(L_{\text{pos,i}} + L_{\text{neg,i}})
    \label{eqn:total-loss}
\end{equation}

The learning process of the network involves calculating the partial derivatives of the total loss function with respect to each weight parameter at the current layer $i$, and this error information is detached from the latter layers in the gradient graph. Before the activations are passed to the next layer, they undergo L2 normalization. This normalization prevents the subsequent layer from learning an identity mapping if the current layer already produces high goodness values for positive samples and low goodness values for negative samples.

To make a prediction, FF tests every possible label separately. For each candidate label, the input is encoded with that label and passed through the network, producing goodness values at each layer. These layer-wise goodness values are summed to form a single score for that label. The network then selects the label with the highest score as the output.

\begin{figure}[H]
    \centering
    \includegraphics[width=0.99\linewidth]{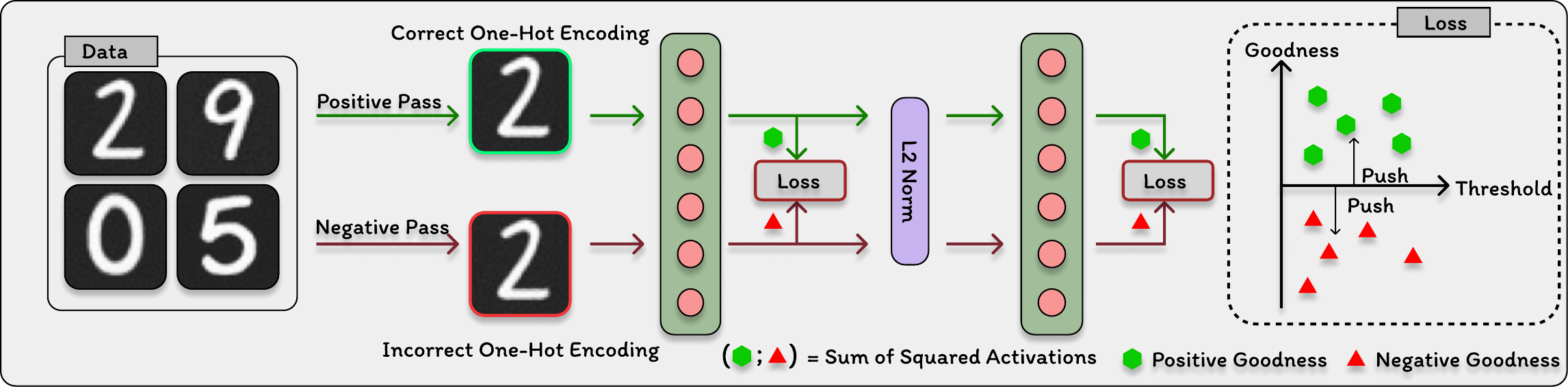}
    \caption{Visualization of the Forward-Forward algorithm.}
    \label{fig:forwardforwardvisual}
\end{figure}


\begin{figure}[H]
    \centering
    \includegraphics[width=0.99\columnwidth]{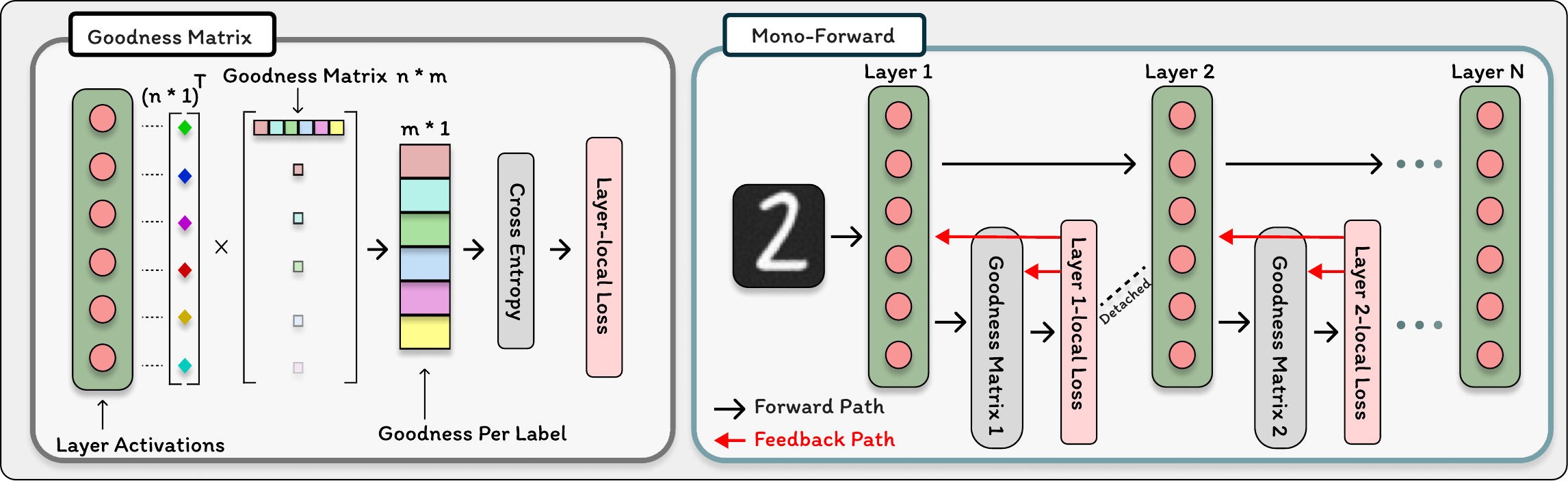}
    \caption{Training architecture of the Mono-Forward algorithm. }
    \label{fig:MF-training-figure}
\end{figure}

In MF, each layer is trained using only its own locally available goodness values, without requiring error signals to be propagated backward from deeper layers. In figure~\ref{fig:MF-training-figure}, the red arrows indicate that the loss value computed at each layer is used to update the parameters in the goodness matrix and the network layer. 



\section{Information Loss in L2 Normalization}
\label{app:l2}

We provide an indirect reconstruction-based proxy suggesting that less input information remains recoverable after inter-layer $L_2$ normalization. Specifically, we train FC-FF + GE models on FashionMNIST using a 5-layer MLP with 500 hidden units per layer, under two conditions: with and without inter-layer $L_2$ normalization. After training, the backbone is frozen, and for each hidden-layer output $z_l$ we train a decoder probe with one hidden layer of 500 neurons to reconstruct the original input image. The primary evaluation metric is the test reconstruction mean squared error (MSE). Lower reconstruction MSE indicates that more input information remains recoverable from the representation.

\begin{figure}[H]
    \centering
    \includegraphics[width=0.6\linewidth]{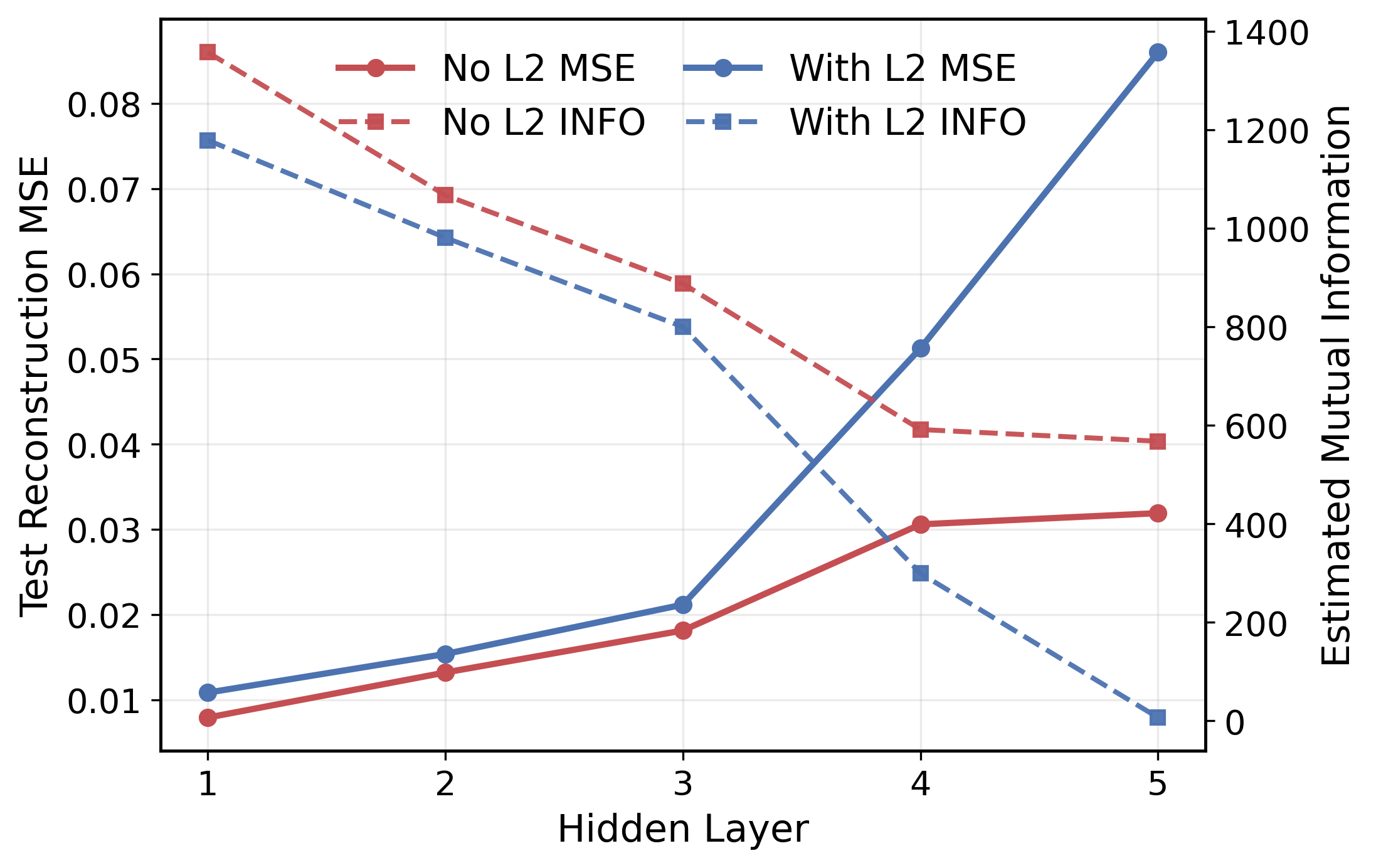}
    \caption{Layer-wise Reconstruction Loss and Estimated Mutual Information. Solid lines show decoder reconstruction MSE (left axis). Dashed lines show a rate-distortion-inspired information proxy (right axis), obtained from the decoder MSE via $\hat I = \frac{d}{2}\log_2(\sigma_X^2/D)$, where $d=784$, $\sigma_X^2$ is the mean input pixel variance, and $D$ is the test reconstruction MSE.}

    \label{fig:information-loss}
\end{figure}

The results in Fig.~\ref{fig:information-loss} demonstrate that the hidden-layer outputs learned without $L_2$ normalization consistently yield lower reconstruction error than those learned with $L_2$ normalization, with the gap increasing in deeper layers. This indicates that introducing $L_2$ normalization between layers causes the network to retain less recoverable information about the input. 

In this experiment, the backbone was trained for 100 epochs using the Adam optimizer with a learning rate of \(0.001\) and a batch size of 128. A step learning rate scheduler was applied with a step size of 15 and a decay factor \(\gamma = 0.1\). The decoder was trained separately for 40 epochs using the Adam optimizer, a learning rate of \(0.001\), a batch size of 512, and an MSE loss objective, without using a learning rate scheduler.

The results for this experiment can be reproduced within 15 minutes using less than 50 MB of memory, under the same setup described in Appendix~\ref{app:mlp-experiments}.

\section{Relation to Prior Methods}
\label{app:relationship-to-prior-methods}
Several existing methods introduce auxiliary objectives at intermediate layers to reduce reliance on global backpropagation, including CaFo \cite{zhao2023cascadedforwardalgorithmneural}, which is also inspired by FF, deeply supervised networks \cite{lee2014deeplysupervisednets}, greedy layerwise training \cite{NIPS2006_5da713a6}, and Decoupled Greedy Learning \cite{belilovsky2020decoupledgreedylearningcnns}.

CaFo \cite{zhao2023cascadedforwardalgorithmneural} is closely related to MF in that it attaches independent projector heads to intermediate convolutional layers to provide layer-specific learning signals. These projector heads allow each layer to compute its own local loss, enabling training without relying solely on a single global objective. In CaFo, the network layers are either not updated or updated using feedback alignment, where fixed random matrices are used to propagate the error signals from the projector heads to the corresponding layers instead of using exact gradient transposes. MF differs from CaFo in how these auxiliary pathways are interpreted and utilized. While CaFo either does not transmit error signals or relies on FA to transmit them to ensure stricter locality, MF allows exact error signals to be transmitted from the goodness matrix to its corresponding network layer. While this softens the claim of locality, this design can be interpreted through the perspective of dendritic computation. In biological neurons, dendritic branches can perform local nonlinear processing and integrate signals independently before influencing the neuron's output \cite{London2005DendriticC}. Under this interpretation, the goodness matrix together with its associated cross-entropy objective can be viewed as analogous to a dendritic branch that generates a local supervisory signal for the corresponding layer.

Deep supervision \cite{lee2014deeplysupervisednets} also introduces auxiliary classifiers at intermediate layers, providing additional supervision signals during training. These auxiliary losses are primarily used to improve optimization and representation learning by encouraging intermediate features to become discriminative earlier in the network. However, deep supervision still relies on standard end-to-end backpropagation to update the network parameters, meaning that gradients from both the auxiliary losses and the final loss propagate through the entire network. Consequently, the auxiliary classifiers in deep supervision act mainly as optimization aids rather than defining independent local learning objectives. In contrast, MF treats each layer as directly responsible for producing class-relevant outputs through its own goodness matrix, allowing each layer to compute a self-contained local objective without requiring a single global loss to coordinate learning across the network.

Greedy layer-wise learning \cite{NIPS2006_5da713a6} is also related to MF in that both avoid relying entirely on end-to-end gradient propagation and instead assign learning responsibilities to individual layers. In classical greedy layer-wise learning, layers are trained sequentially with many local iterations until convergence, typically in an unsupervised manner, with each layer optimized to produce a useful representation before being stacked onto the next. After this stage-wise pretraining process, the full network is often fine-tuned using backpropagation for the final supervised task. MF differs from this paradigm in the way layer updates are scheduled. Instead of fully training each layer to convergence before progressing deeper, MF updates each layer only for one local iteration before the training process proceeds to the next layer, well before the current layer has converged. We demonstrate that increasing the number of iterations is not beneficial in MF in Section ~\ref{experiments}.

Decoupled Greedy Learning (DGL) \cite{belilovsky2020decoupledgreedylearningcnns} similarly reduces reliance on end-to-end gradient propagation by introducing local objectives at intermediate stages of a network. In DGL, the network is partitioned into blocks, and auxiliary networks are attached to these blocks to compute local losses, enabling each block to be trained independently and potentially in parallel. However, DGL is primarily designed to decouple the training process across network blocks to enable parallel and asynchronous updates, relying on mechanisms such as replay buffers to stabilize training. Moreover, because DGL partitions the network into relatively large blocks containing multiple layers, the resulting learning signals are less strictly local. In contrast, MF attaches a lightweight goodness matrix to each individual layer and optimizes it using a local classification objective derived from the decomposition of the Forward-Forward algorithm.

\section{Relation to Biological Learning Principles}
\label{app:biological-plausibility}
Several aspects of MF are consistent with biological learning principles. First, MF preserves the use of layer-local learning signals, where parameter updates depend only on locally available information from the corresponding layer and its goodness matrix. This aligns with the widely accepted view that biological synaptic plasticity is governed by local activity signals rather than globally propagated gradients \cite{Song_Lukasiewicz_Xu_Bogacz_2020}.

Second, the goodness matrices used in MF can be interpreted through the perspective of dendritic computation. Biological neurons contain dendritic branches that perform local nonlinear processing and can generate intermediate signals before influencing the neuron’s output \cite{London2005DendriticC}. Under this interpretation, the goodness matrix together with its associated classification objective may be viewed as analogous to a dendritic branch that produces a local supervisory signal for the corresponding layer.

Third, cortical learning is believed to involve multiple feedback pathways that provide supervisory signals to different layers or regions of the brain rather than relying on a single global error signal \cite{cell}. The layer-specific goodness matrices in MF can be interpreted as providing such localized supervisory feedback that guides representation learning throughout the network.

Lastly, we emphasize that MF should not be viewed as a faithful biological model. Instead, its learning mechanism can be understood as inspired by biological learning principles, particularly the use of local learning signals and separate supervision across layers. At the same time, MF still relies on explicit labels and gradient-based optimization, which are unlikely to directly correspond to biological learning mechanisms.

\section{MLPs Experimental Setup}
\label{app:mlp-experiments}

For MNIST and FashionMNIST, we used a $2\times1000$ ReLU-activated multilayer perceptron (MLP), while for CIFAR-10 and CIFAR-100, a deeper $3\times2000$ ReLU MLP was employed. These architectures were selected not to maximize performance, but to ensure a fair comparison of the algorithms under identical experimental conditions. Key hyperparameters, including the learning rate and step size, were fine-tuned prior to reporting the results.

For all datasets, inputs were normalized using dataset-specific channel statistics and then flattened before being passed to the MLP. Specifically, MNIST used mean/std $(0.1307, 0.3081)$, FashionMNIST used $(0.2860, 0.3530)$, CIFAR-10 used mean $(0.4914, 0.4822, 0.4465)$ and std $(0.2471, 0.2435, 0.2616)$, and CIFAR-100 used mean $(0.5071, 0.4865, 0.4409)$ and std $(0.2673, 0.2564, 0.2762)$.

\begin{table}[H]
\centering
\small
\setlength{\tabcolsep}{4pt}
\renewcommand{\arraystretch}{1.08}
\caption{Hyperparameter search space for the MLP experiments.}
\label{tab:mlp_grid_search}
\begin{tabular}{@{}p{0.30\linewidth}p{0.62\linewidth}@{}}
\toprule
Hyperparameter & Values / setting \\
\midrule
Datasets & MNIST, FashionMNIST, CIFAR-10, CIFAR-100 \\
Architecture (MNIST Family) & $2\times1000$ ReLU MLP \\
Architecture (CIFAR Family) & $3\times2000$ ReLU MLP \\
Optimizer & \{SGD, Adam\} \\
Learning rate & $\{0.2, 0.1, 0.02, 0.01, 0.005, 0.001\}$ \\
Scheduler & StepLR with decay factor $\gamma=0.1$ \\
Scheduler step size & $\{15, 20, 30\}$ epochs \\
Training epochs & 200 \\
Batch size & $\{64, 128, 256\}$ \\
\bottomrule
\end{tabular}
\end{table}

All models were trained for 200 epochs without dropout, weight decay, or data augmentation to ensure we do not claim any advantages from experimental setup. Hyperparameters, as shown in Table~\ref{tab:mlp_grid_search}, were selected by grid search over the optimizer, learning rate, scheduler step size, and batch size, while keeping the architecture and training duration fixed. We used a StepLR scheduler with decay factor $\gamma=0.1$ in all runs. The best configuration for each method and dataset was then used to report the final test performance. Results in Table~\ref{tab:accuracy_comparison_mlp} are averaged over 3 random seeds. Reproduced results are marked with $^*$.

\paragraph{System setup.} 
All measurements were collected on a single machine with one NVIDIA GeForce RTX 4090 GPU (23.5\,GB VRAM), 32 logical CPU cores, and 62.5\,GB system memory, running Linux with Python 3.10 and PyTorch 2.6. 

To estimate total compute, we measured the first 10 epochs of each configuration and extrapolated to the full training horizon (200 epochs for the MLP experiments). The MLP experiments yielded an estimated 256.3 GPU-hours for hyperparameter search and 7.1 GPU-hours for the final 3-seed evaluations, for a total of 263.5 GPU-hours (11.0 GPU-days), with a minimum memory requirement of 4 GB.

\section{CNNs Experimental Setup}
\label{app:cnn-setup}
The reproduced CNN results were obtained using a four-block convolutional network with channel widths of 64, 128, 256, and 512. Each block consisted of a \(3 \times 3\) convolution, batch normalization, a ReLU activation, and \(2 \times 2\) max-pooling. For the BP baseline, the final feature map was passed through global average pooling followed by a linear classifier. Batch normalization \cite{batchnorm} was retained in the BP model to provide a stronger supervised baseline. For MF, a local linear head was attached after each convolutional layer. All inputs were normalized using the same dataset-specific normalization procedure described in Appendix~\ref{app:mlp-experiments}. As in the MLP experiments, we performed a grid search over key hyperparameters, including learning rate, batch size, and optimizer. The explored hyperparameter space is reported in Table~\ref{tab:cnn_grid_search}.

For the 4-layer CNN experiments, the procedure yielded 40.4 GPU-hours for hyperparameter search and 3.4 GPU-hours for the final 3-seed evaluations, for a total of 43.8 GPU-hours (1.82 GPU-days), with the same system setup in Appendix~\ref{app:mlp-experiments} and a minimum memory requirement of 2 GB.

\begin{table}[t]
\centering
\small
\setlength{\tabcolsep}{4pt}
\renewcommand{\arraystretch}{1.08}
\caption{Hyperparameter space explored for the CNN experiments.}
\label{tab:cnn_grid_search}
\begin{tabular}{@{}p{0.30\linewidth}p{0.62\linewidth}@{}}
\toprule
Hyperparameter & Values / setting \\
\midrule
Architecture & Conv-BN-ReLU-MaxPool CNN \\
Channels & 64/128/256/512 \\
Optimizer & \{SGD, Adam\} \\
Learning rate & $\{0.2, 0.1, 0.02, 0.01, 0.005, 0.001\}$ \\
Scheduler & Cosine Annealing \\
Training epochs & 300 \\
Batch size & \{64, 128, 256\} \\
\bottomrule
\end{tabular}
\end{table}

\section{Representation Quality Analysis}
\label{app:representation-quality}

\begin{figure*}
    \centering
    \includegraphics[width=0.99\linewidth]{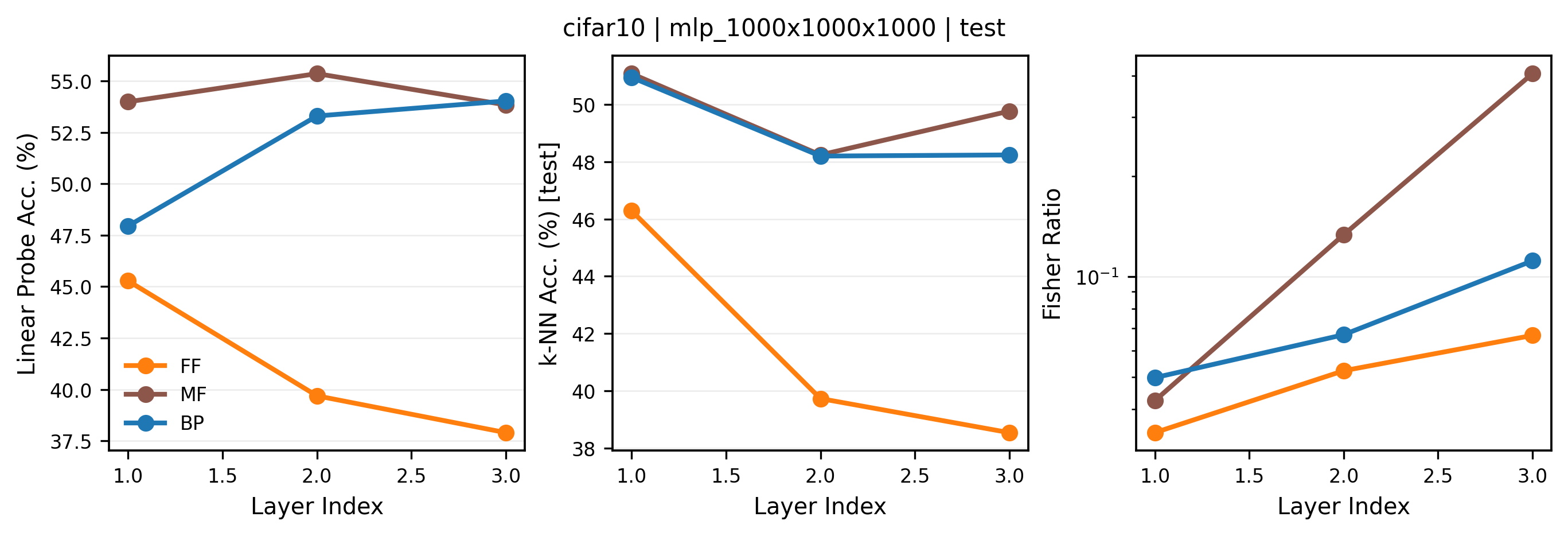}
    \caption{Layer-wise representation quality on CIFAR-10 using a 1000×1000×1000 MLP.}
    \label{fig:representation}
\end{figure*}
Linear-probe accuracy measures how easily class labels can be decoded from the frozen features of a layer using a trained linear classifier. k-NN accuracy provides a non-parametric view of representation quality by classifying each test feature according to the majority label among its $k=10$ nearest training features in the same representation space. Formally, for a test representation $z$, its predicted label is defined as
\[
\hat{y}(z)=\operatorname{mode}\{y_j : z_j \in \mathcal{N}_k(z)\},
\]
where \(\mathcal{N}_k(z)\) denotes the set of the \(k\) nearest training representations to \(z\), and \(\operatorname{mode}(\cdot)\) returns the most frequent label among them. Fisher ratio measures class separability through the ratio of between-class to within-class scatter,
\[
\mathrm{Fisher\ Ratio}=\frac{\operatorname{tr}(S_B)}{\operatorname{tr}(S_W)},
\]
where
\[
S_B=\sum_{c=1}^{C} n_c (\mu_c-\mu)(\mu_c-\mu)^\top,
\]
\[
S_W=\sum_{c=1}^{C}\sum_{i:y_i=c}(z_i-\mu_c)(z_i-\mu_c)^\top.
\]

Here, \(C\) is the number of classes, \(n_c\) is the number of samples in class \(c\), \(\mu_c\) is the mean feature vector of class \(c\), and \(\mu\) is the global mean feature vector over all samples. The notation \(\operatorname{tr}(\cdot)\) denotes the trace of a matrix, i.e., the sum of its diagonal elements.

On CIFAR-10 with the 1000×1000×1000 MLP summarized in Fig.~\ref{fig:representation}, MF consistently outperforms FF across all layers on all three measures. Although linear-probe accuracy is partly aligned with MF’s local classification objective, the same overall trend is also observed for k-NN accuracy and Fisher ratio, which provide non-parametric and geometric views of representation quality. In particular, while FF exhibits progressively weaker class structure in deeper layers, MF maintains strong layer-wise decodability and substantially stronger class separation, especially in the deeper layers where its Fisher ratio rises markedly above both FF and BP. MF is also competitive with, and in some layers slightly exceeds, BP on linear-probe and k-NN evaluation.

\section{Convergence Behavior of MF}
\label{app:convergence-behavior}
\begin{figure*}[t]
    \centering
    \includegraphics[width=0.99\linewidth]{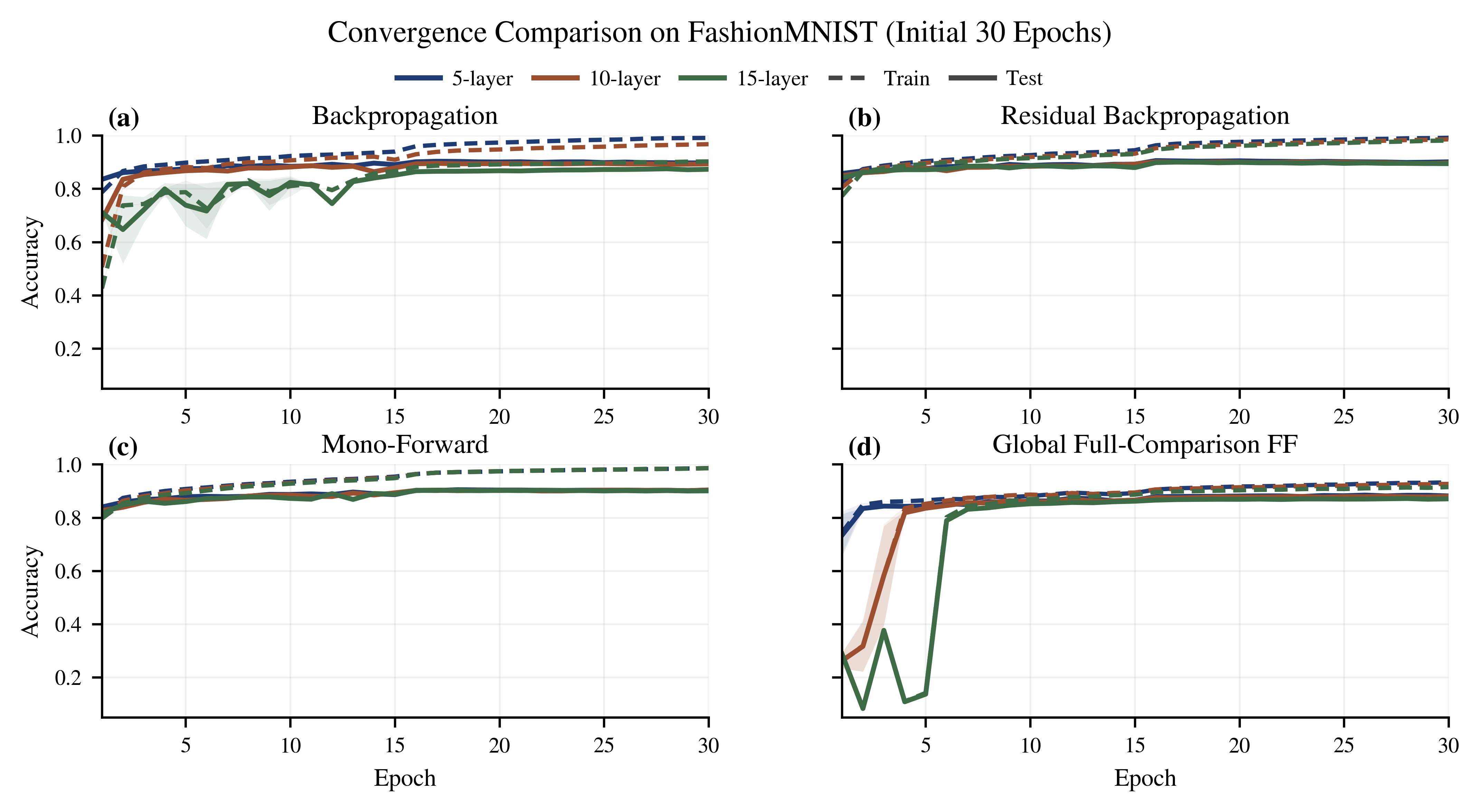}
    \caption{The convergence behavior of BP, BP with Residual Connection, MF, and (FC-FF + GE).}
    \label{fig:convergence-comparison}
\end{figure*}

To evaluate the convergence behavior of MF compared to BP with increasing network depth, we trained MLPs comprising 5, 10, and 15 layers, each containing 1000 neurons with Adam optimizer, batch size of 256, and a fixed learning rate of 0.001. It is well-established that standard BP encounters gradient instability issues as network depth increases; therefore, we have also included a residual connection variant of BP to address this known limitation. For context, we have also included (FC-FF + GE), as the convergence behavior of vanilla FF performs substantially worse. Our results, presented in Fig. \ref{fig:convergence-comparison}, illustrate that vanilla BP and (FC-FF + GE) exhibit significant convergence instability as the network depth increases. In contrast, both MF and residual BP demonstrate notably more stable convergence behaviors.

The time and memory required to collect these results, with the same system setup in Appendix~\ref{app:mlp-experiments}, are presented in Table~\ref{tab:convergence_compute}.

\begin{table}[t]
\centering
\small
\setlength{\tabcolsep}{6pt}
\renewcommand{\arraystretch}{1.08}
\caption{Estimated compute and peak GPU memory for the FashionMNIST convergence experiment. Times are extrapolated from 10 measured epochs to 30 epochs.}
\label{tab:convergence_compute}
\begin{tabular}{llcc}
\toprule
Method & Depth & Est. wall time (s) & Peak memory (MB) \\
\midrule
BP & 5  & 99.1  & 111.1 \\
BP & 10 & 99.4  & 210.4 \\
BP & 15 & 106.3 & 310.5 \\
Residual BP & 5  & 97.2  & 112.0 \\
Residual BP & 10 & 100.5 & 210.4 \\
Residual BP & 15 & 108.8 & 310.5 \\
MF & 5  & 108.2 & 99.3 \\
MF & 10 & 124.3 & 180.2 \\
MF & 15 & 137.5 & 261.0 \\
FC-FF + GE & 5  & 189.9 & 210.2 \\
FC-FF + GE & 10 & 218.4 & 390.1 \\
FC-FF + GE & 15 & 249.1 & 570.3 \\
\bottomrule
\end{tabular}
\end{table}

\section{Ablation Study on Local Iterations}
\label{app:local-iter}
To analyze the effect of local iterations, we implement MF on a VGG-16 architecture \cite{simonyan2015deepconvolutionalnetworkslargescale} trained on CIFAR-10, attaching a goodness matrix to the output of each convolutional layer. We vary the number of local iteration  $K \in \{1,3,5\}$ performed for each layer before propagating its activations forward. In MF, each layer is optimized using an independent local objective. As a result, deeper layers are trained on representations produced by earlier layers whose parameters are simultaneously evolving. If these upstream representations change significantly during training, the downstream optimization problem becomes non-stationary, potentially destabilizing learning. To quantify this effect, we measure the representation drift observed at the input of the final convolutional layer over the first 20 training epochs using a fixed probe batch. The representation drift $D^{(t)}$ is defined as the normalized change in the activation tensor between consecutive epochs:
\begin{equation}
D^{(t)} =
\frac{\left\| h^{(t)} - h^{(t-1)} \right\|_2}
{\left\| h^{(t-1)} \right\|_2},
\end{equation}
where $h^{(t)}$ denotes the activation at the input of the final convolutional layer computed on the fixed probe batch after epoch $t$, and $\|\cdot\|_2$ denotes the Euclidean norm. 

\begin{figure*}[h]
\centering

\begin{minipage}{0.32\textwidth}
\centering
\includegraphics[width=\linewidth]{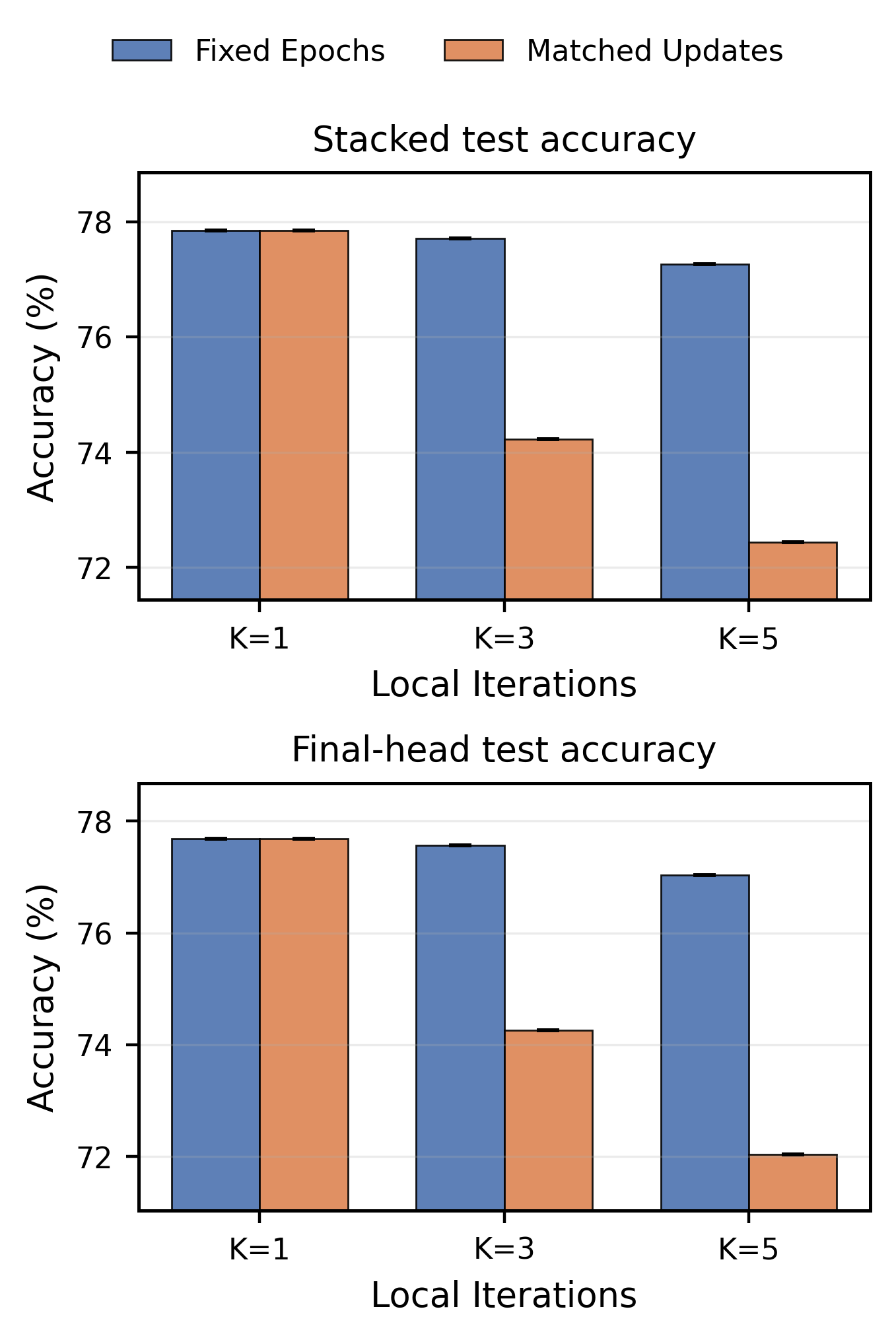}
\end{minipage}
\hfill
\begin{minipage}{0.32\textwidth}
\centering
\includegraphics[width=\linewidth]{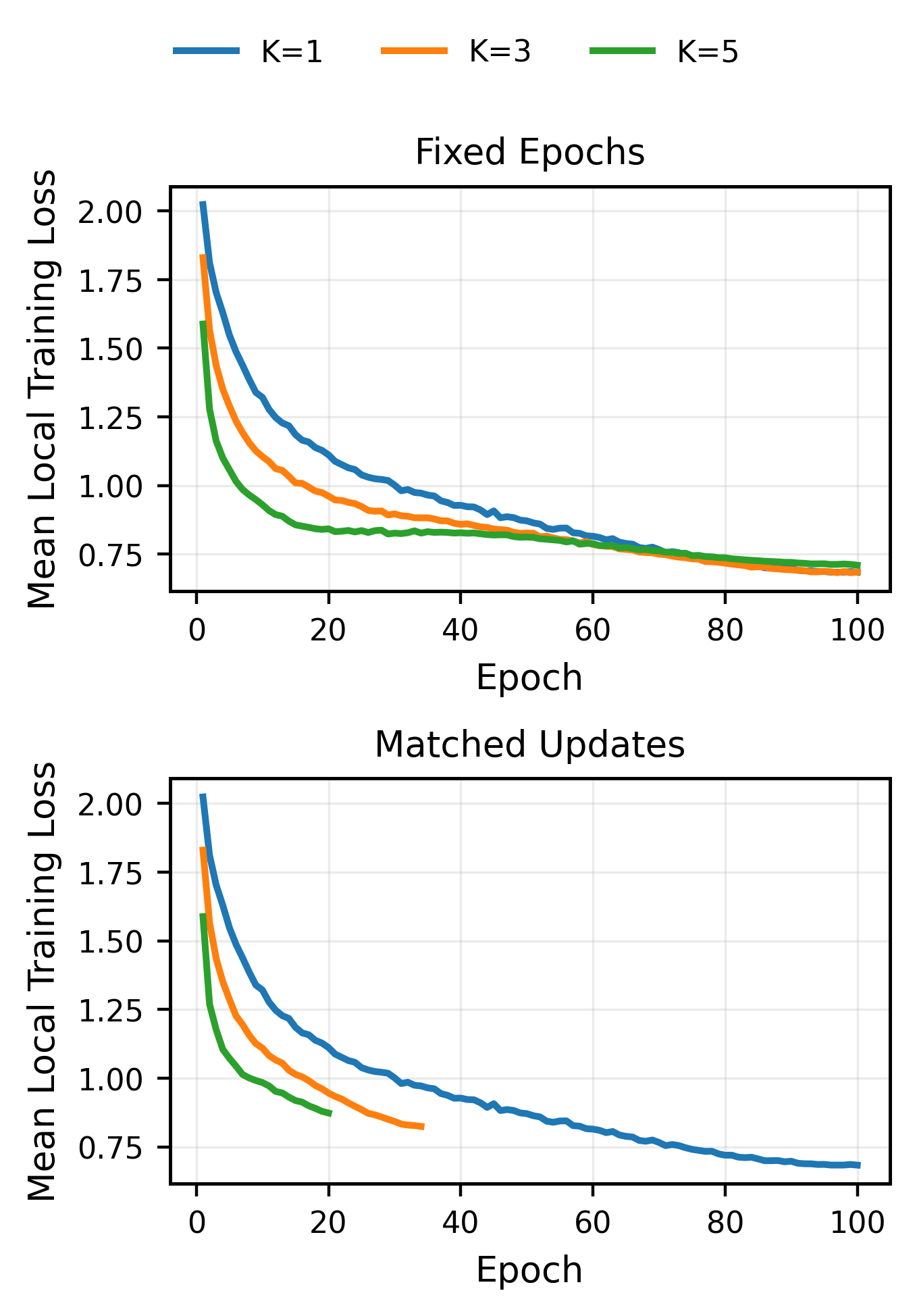}
\end{minipage}
\hfill
\begin{minipage}{0.32\textwidth}
\centering
\includegraphics[width=\linewidth]{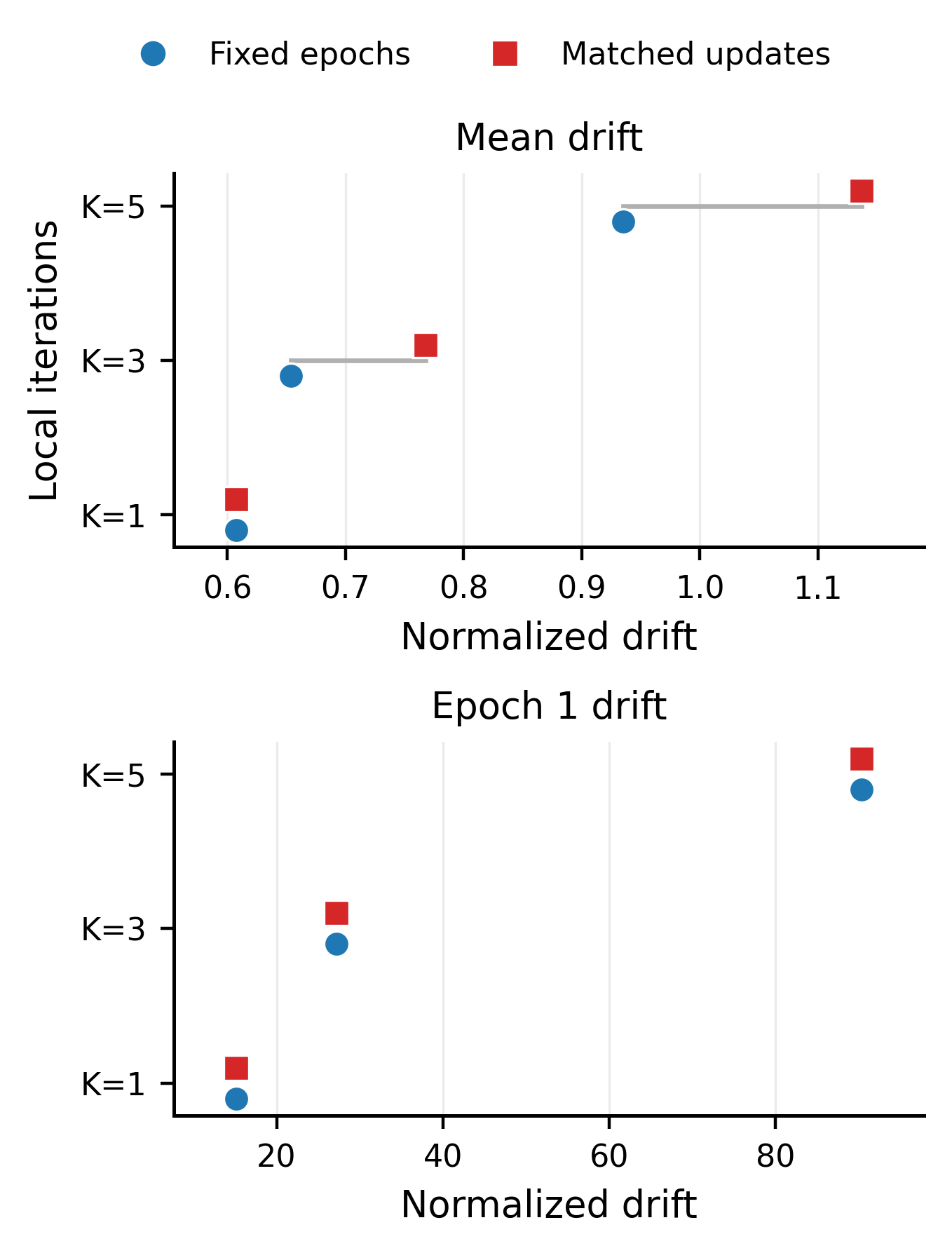}
\end{minipage}

\caption{Training dynamics of MF with different local iteration counts $K \in \{1,3,5\}$ on VGG-16 trained on CIFAR-10.}
\label{fig:drift_comparison}

\end{figure*}

We evaluate the effect of increasing the number of local optimization steps per layer using two training protocols, namely fixed-epoch setting and matched-update setting. In the fixed-epoch setting, all models are trained for 100 epochs regardless of the local iteration count $K$. In the matched-update setting, we approximately equalize the total number of local optimization steps  across configurations by scaling the number of training epochs inversely with $K$. For instance, the number of epochs is set to 100, 34, and 20 for $K = {1,3,5}$, respectively, so that each configuration performs a comparable number of local updates. 

The results are summarized in Fig.~\ref{fig:drift_comparison}. As shown in Fig.~\ref{fig:drift_comparison} (right), increasing the number of local iterations consistently leads to larger representation drift in both training protocols, indicating that the representations received by the final layer become more unstable as $K$ increases. For completeness, we also report the corresponding test accuracy and training loss curves in Fig.~\ref{fig:drift_comparison} (left and middle, respectively). These results show that increasing $K$ does not yield any improvements in predictive performance while introducing greater representation instability.

Table~\ref{tab:vgg16-mf-local-iters} reports the wall time and peak memory needed to reproduce this experiment under the hardware setup described in Appendix~\ref{app:mlp-experiments}. Overall, the full experiment can be reproduced in under two GPU-hours.

\begin{table}[t]
\centering
\small
\caption{VGG16 MF on CIFAR-10 with a local head on each conv layer, using Adam and batch size 128 on an RTX 4090. Wall time is measured for 1 epoch and extrapolated to 20 epochs. Peak memory is peak training GPU memory.}
\label{tab:vgg16-mf-local-iters}
\begin{tabular}{rcc}
\toprule
Local iters & Total wall time (s) & Peak memory (MB) \\
\midrule
1 & 81.36 & 362.13 \\
3  & 180.15 & 362.75 \\
5 & 292.63 & 362.75 \\
\bottomrule
\end{tabular}
\end{table}

\section{Memory Consumption of MF}
\label{app:memory-consumption}

\begin{figure*}[h]
    \centering
    \includegraphics[width=1\columnwidth]{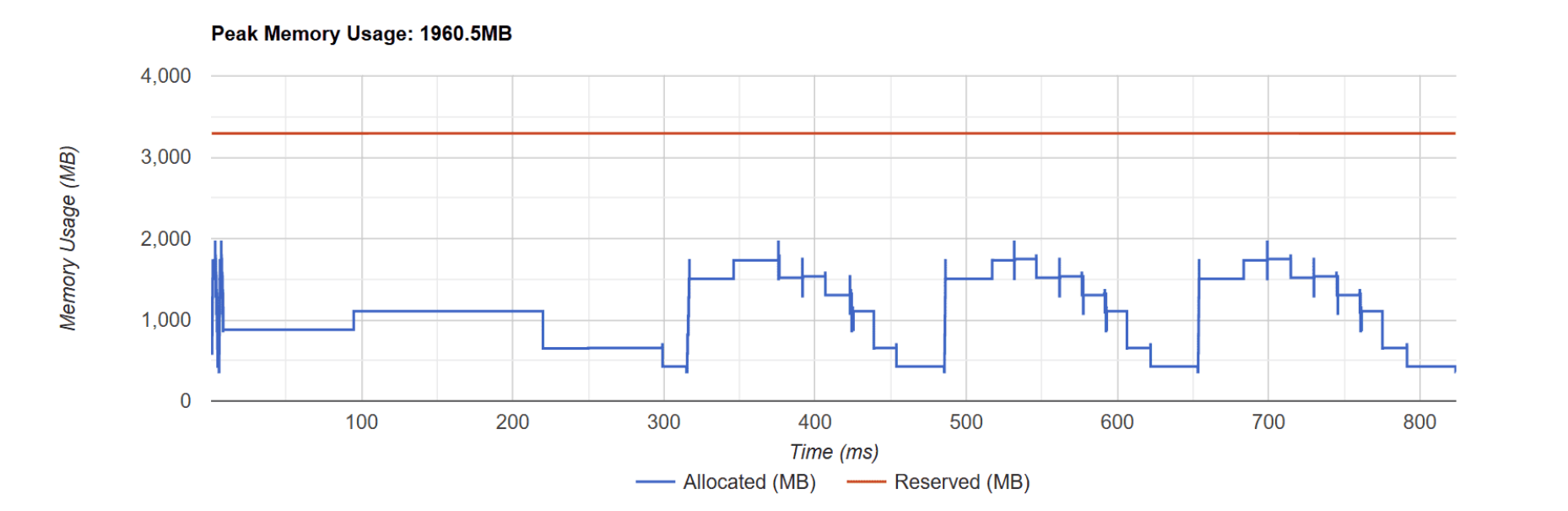}
    \caption{Memory Consumed during Training under BP. This experiment utilizes MNIST dataset on a network of size $5 \times 2000$ with batch size 30000 and SGD optimizer.}
    \label{fig:BP-memory 30000}
\end{figure*}

\begin{figure*}[h]
    \centering
    \includegraphics[width=1\columnwidth]{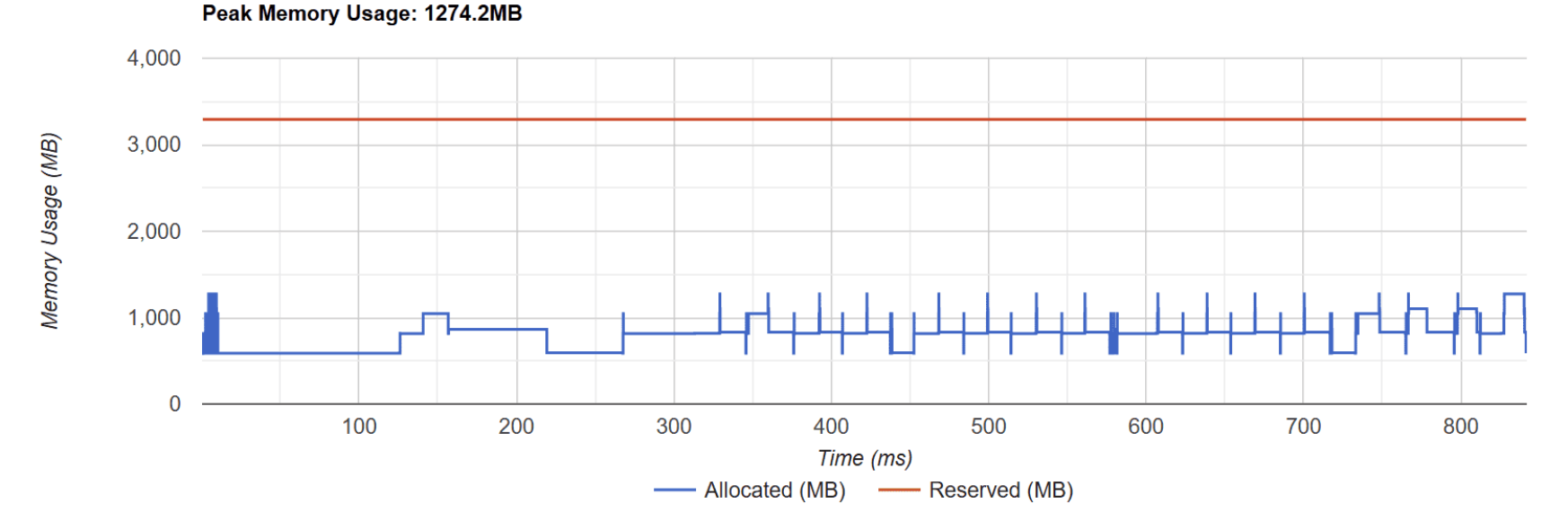}
    \caption{Memory Consumed during Training under MF. This experiment uses the MNIST dataset on a network of size 5 * 2000 with batch size 30000 and SGD optimizer.}
    \label{fig:MF-memory 30000}
\end{figure*}

We analyze the memory consumption during training for both MF and BP, as shown in Fig. \ref{fig:BP-memory 30000} and \ref{fig:MF-memory 30000}. The results confirm that MF requires less memory.

\section{CNN Limitations in MF}
\label{app:cnn-limitation}
CNN architectures typically reduce spatial resolution as the network deepens through pooling or strided convolutions. This progressively shrinks the feature maps, which helps control the computational cost. However, the MF framework attaches a goodness matrix to each layer to map the layer activations to class goodness scores. When this operation is applied to early convolutional layers, the feature maps still have large spatial dimensions (e.g., many pixels per channel). As a result, the activation tensor contains a very large number of values. Applying a goodness matrix to these large tensors requires significantly more memory than in later layers where spatial dimensions have already been reduced. This issue becomes even more pronounced when the number of classes is large, because the goodness matrix must produce a goodness score for each class. Consequently, the memory cost scales with both the spatial size of the feature maps and the number of classes.

\begin{figure}[h]
    \centering
    \includegraphics[width=0.99\linewidth]{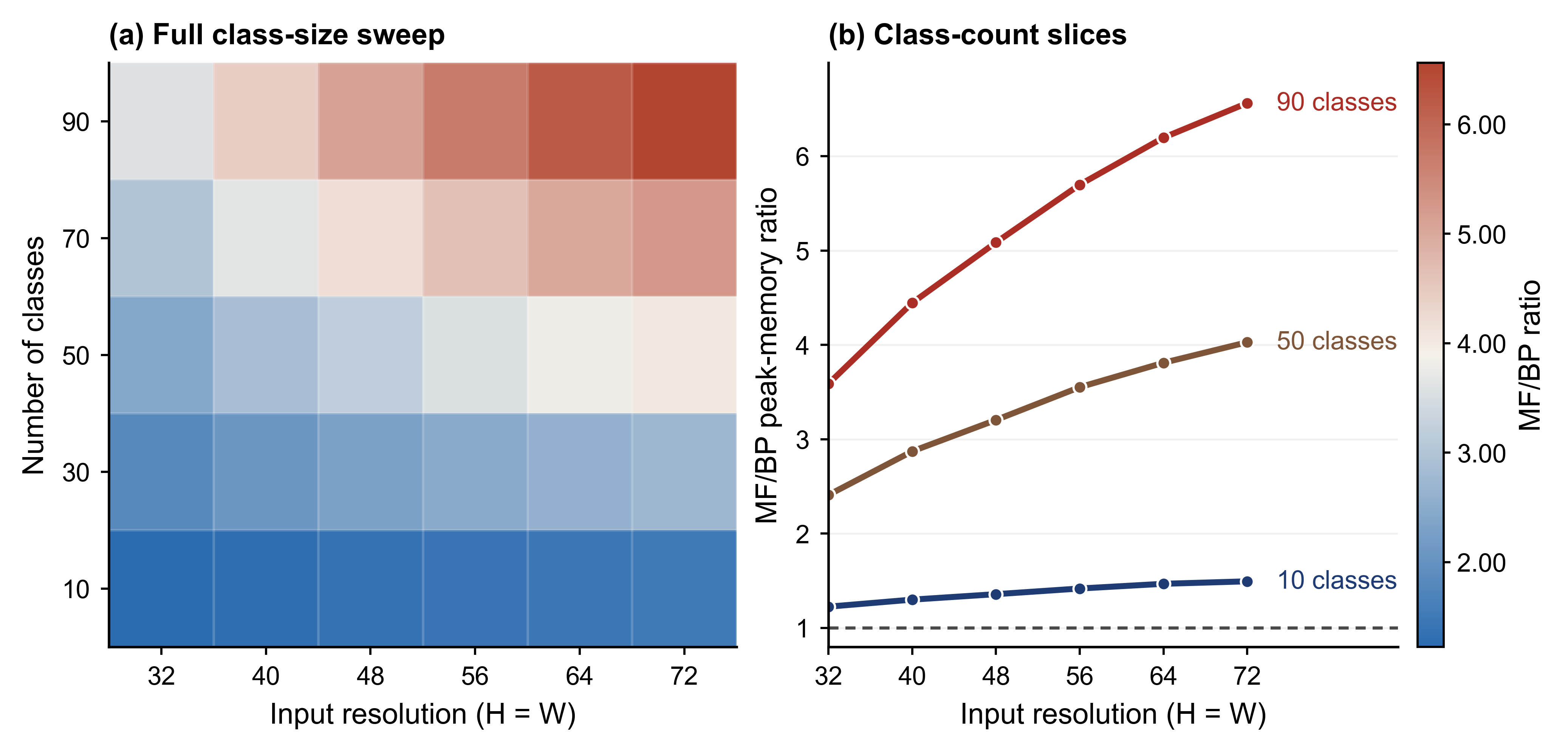}
    \caption{MF/BP memory ratio in ResNet-18. The dashed horizontal line (solid black line in the heatmap) marks where the memory consumption is the same for BP and MF. The colored dots indicate the number of classes present. This experiment is conducted with a batch size of 32 on ResNet-18 with SGD optimizer.}
    \label{fig:class-sweep}
\end{figure}

To investigate this effect, we conduct a sweep in which the number of classes is increased from 10 to 90 in increments of 20, while the input image resolution is increased from 32×32 to 72×72 in increments of 8×8. The results, summarized in Fig.~\ref{fig:class-sweep}, show that the memory consumption of MF grows rapidly with both spatial resolution and class count. In particular, under the ResNet-18 architecture, the memory usage of MF exceeds that of standard BP training, with the gap growing significantly as both resolution and class count increase. This highlights that MF may not retain the memory advantage observed in MLP architectures.

The experimental results were obtained using a full sweep time of approximately 45 seconds, following the same setup described in Appendix~\ref{app:mlp-experiments}. Under this setting, the peak memory usage was 514 MB for BP and 3354 MB for MF.

\section{MF in MLP-Mixers}
\label{app:mlp-mixer}

We experiment with six different MLP-Mixer configurations across five datasets: CIFAR-10, CIFAR-100, COIL-100 \cite{nene96columbia}, PathMNIST \cite{Yang_2023}, and TinyImageNet \cite{5206848}. The results are summarized in Table~\ref{tab:mixer_single_table_bp_vs_mf}. While BP achieves higher test accuracy on most tasks, MF consistently requires no more than 42\% of the memory used by BP. Notably, on PathMNIST, MF slightly outperforms BP by 0.6\% (best in MF minus best in BP) while requiring only 31\% of the memory consumed by BP. 

    The favorable results on PathMNIST motivate a broader question: whether the same behaviour can also be observed on datasets with similar low-resolution medical-image characteristics. Since MF slightly outperforms BP on PathMNIST while using substantially less memory, we extend the evaluation to additional MedMNIST datasets. In this extended sweep, we report both accuracy and AUC to capture classification performance more comprehensively. We find that MF achieves higher accuracy than BP in 33\% of the configurations and higher AUC in 20\% of the configurations. MF mainly preserves its memory-efficiency advantage, while BP generally remains stronger or comparable in predictive performance. The results of this extended MedMNIST evaluation are reported in Table~\ref{tab:medmnist_mixer_bp_vs_mf}.
\begin{table*}[h]
\centering
\footnotesize
\setlength{\tabcolsep}{4.0pt}
\renewcommand{\arraystretch}{1.12}
\caption{Unified architecture sweep of MLP-Mixer across datasets. Each row is one architecture (depth $\times$ width), e.g. D5-W256 contains 5 mixer layers with dimension 256 neurons. Reported values are means over 3 runs. $\Delta$ is MF minus BP.}
\vspace{7pt}
\label{tab:mixer_single_table_bp_vs_mf}
\resizebox{\textwidth}{!}{%
\begin{tabular}{llrrrrrrrrr}
\toprule
Dataset & Arch & BP Top-1 & MF Top-1 & $\Delta$Top-1 & BP Top-5 & MF Top-5 & $\Delta$Top-5 & BP Mem (GB) & MF Mem (GB) & MF/BP Mem \\
\midrule
CIFAR-10 & D5-W256 & 81.66 & \textbf{78.60} & -3.06 & 98.12 & 98.46 & 0.34 & 0.629 & 0.238 & 0.38 \\
CIFAR-10 & D5-W512 & 81.00 & 77.58 & -3.42 & \textbf{98.67} & 98.37 & -0.30 & 1.682 & 0.596 & 0.35 \\
CIFAR-10 & D8-W256 & \textbf{81.76} & 78.48 & -3.28 & 98.43 & \textbf{98.52} & 0.09 & 0.968 & 0.263 & 0.27 \\
CIFAR-10 & D8-W512 & 81.60 & 77.65 & -3.95 & 98.13 & 98.25 & 0.12 & 2.635 & 0.694 & 0.26 \\
CIFAR-10 & D12-W256 & 80.88 & 77.52 & -3.36 & 97.92 & 98.25 & 0.33 & 1.419 & 0.296 & 0.21 \\
CIFAR-10 & D12-W512 & 80.67 & 76.65 & -4.02 & 98.45 & 98.00 & -0.45 & 3.904 & 0.824 & 0.21 \\
\midrule
CIFAR-100 & D5-W256 & 52.19 & 46.14 & -6.05 & 80.08 & 74.13 & -5.95 & 0.629 & 0.240 & 0.38 \\
CIFAR-100 & D5-W512 & 52.31 & \textbf{46.75} & -5.56 & \textbf{80.64} & \textbf{76.40} & -4.24 & 1.683 & 0.600 & 0.36 \\
CIFAR-100 & D8-W256 & 51.27 & 46.62 & -4.65 & 79.57 & 73.43 & -6.14 & 0.968 & 0.266 & 0.27 \\
CIFAR-100 & D8-W512 & 51.82 & 46.28 & -5.54 & 80.25 & 74.54 & -5.71 & 2.635 & 0.699 & 0.27 \\
CIFAR-100 & D12-W256 & \textbf{52.68} & 46.49 & -6.19 & 79.71 & 72.38 & -7.33 & 1.420 & 0.301 & 0.21 \\
CIFAR-100 & D12-W512 & 52.13 & 46.12 & -6.01 & 80.41 & 74.15 & -6.26 & 3.905 & 0.832 & 0.21 \\
\midrule
COIL-100 & D5-W256 & 99.80 & \textbf{99.40} & -0.40 & 100.00 & 99.97 & -0.03 & 0.664 & 0.263 & 0.40 \\
COIL-100 & D5-W512 & 99.87 & 98.93 & -0.93 & 100.00 & 99.97 & -0.03 & 1.722 & 0.628 & 0.36 \\
COIL-100 & D8-W256 & 99.53 & 99.20 & -0.33 & \textbf{100.00} & \textbf{100.00} & 0.00 & 1.003 & 0.289 & 0.29 \\
COIL-100 & D8-W512 & 99.43 & 99.20 & -0.23 & 100.00 & 100.00 & 0.00 & 2.674 & 0.726 & 0.27 \\
COIL-100 & D12-W256 & \textbf{99.90} & 99.23 & -0.67 & 100.00 & 99.97 & -0.03 & 1.455 & 0.324 & 0.22 \\
COIL-100 & D12-W512 & 99.70 & 98.77 & -0.93 & 100.00 & 99.70 & -0.30 & 3.944 & 0.858 & 0.22 \\
\midrule
PathMNIST & D5-W256 & 87.62 & 89.32 & 1.71 & \textbf{99.51} & \textbf{99.84} & 0.33 & 0.359 & 0.148 & 0.41 \\
PathMNIST & D5-W512 & 88.86 & 88.56 & -0.30 & 99.49 & 99.75 & 0.26 & 1.148 & 0.478 & 0.42 \\
PathMNIST & D8-W256 & 88.63 & \textbf{89.49} & 0.86 & 99.23 & 99.76 & 0.53 & 0.547 & 0.171 & 0.31 \\
PathMNIST & D8-W512 & \textbf{88.89} & 88.51 & -0.38 & 99.51 & 99.76 & 0.25 & 1.801 & 0.573 & 0.32 \\
PathMNIST & D12-W256 & 88.22 & 88.81 & 0.59 & 99.45 & 99.74 & 0.29 & 0.798 & 0.204 & 0.26 \\
PathMNIST & D12-W512 & 87.49 & 88.72 & 1.23 & 99.44 & 99.62 & 0.19 & 2.668 & 0.700 & 0.26 \\
\midrule
TinyImageNet & D5-W256 & \textbf{38.94} & 31.06 & -7.88 & 64.82 & 55.39 & -9.43 & 0.637 & 0.246 & 0.39 \\
TinyImageNet & D5-W512 & 38.60 & \textbf{32.86} & -5.74 & 64.02 & 58.51 & -5.51 & 1.692 & 0.608 & 0.36 \\
TinyImageNet & D8-W256 & 38.68 & 29.89 & -8.79 & 65.31 & 53.17 & -12.14 & 0.976 & 0.273 & 0.28 \\
TinyImageNet & D8-W512 & 38.58 & 32.40 & -6.18 & 65.50 & \textbf{57.80} & -7.70 & 2.644 & 0.710 & 0.27 \\
TinyImageNet & D12-W256 & 38.21 & 30.74 & -7.47 & \textbf{65.51} & 55.93 & -9.58 & 1.427 & 0.310 & 0.22 \\
TinyImageNet & D12-W512 & 38.01 & 30.59 & -7.42 & 64.33 & 55.03 & -9.30 & 3.914 & 0.845 & 0.22 \\
\bottomrule
\end{tabular}%
}
\end{table*}

\begin{table*}[t]
\centering
\footnotesize
\setlength{\tabcolsep}{4.0pt}
\renewcommand{\arraystretch}{1.12}
\caption{Unified architecture sweep of MLP-Mixer across MedMNIST datasets. Reported values are means over 3 runs.}
\vspace{7pt}
\label{tab:medmnist_mixer_bp_vs_mf}
\resizebox{\textwidth}{!}{%
\begin{tabular}{llrrrrrrrrr}
\toprule
Dataset & Arch & BP Acc & MF Acc & $\Delta$Acc & BP AUC & MF AUC & $\Delta$AUC & BP Mem (GB) & MF Mem (GB) & MF/BP Mem \\
\midrule
ChestMNIST & D5-W256 & 94.76 & \textbf{94.76} & -0.00 & 72.23 & 71.03 & -1.20 & 0.357 & 0.145 & 0.41 \\
ChestMNIST & D5-W512 & 94.76 & 94.75 & -0.01 & 72.72 & 71.79 & -0.93 & 1.146 & 0.476 & 0.42 \\
ChestMNIST & D8-W256 & 94.76 & 94.75 & -0.00 & 72.11 & \textbf{72.29} & 0.18 & 0.545 & 0.170 & 0.31 \\
ChestMNIST & D8-W512 & \textbf{94.76} & 94.75 & -0.02 & 72.32 & 71.93 & -0.40 & 1.797 & 0.571 & 0.32 \\
ChestMNIST & D12-W256 & 94.75 & 94.75 & 0.00 & \textbf{72.77} & 71.00 & -1.77 & 0.796 & 0.202 & 0.25 \\
ChestMNIST & D12-W512 & 94.75 & 94.75 & -0.00 & 70.74 & 71.31 & 0.57 & 2.665 & 0.698 & 0.26 \\
\midrule
DermaMNIST & D5-W256 & 75.51 & 74.86 & -0.65 & 90.03 & 91.01 & 0.98 & 0.359 & 0.147 & 0.41 \\
DermaMNIST & D5-W512 & 76.01 & 74.31 & -1.70 & 91.23 & 87.31 & -3.93 & 1.148 & 0.478 & 0.42 \\
DermaMNIST & D8-W256 & 75.46 & \textbf{75.46} & 0.00 & \textbf{92.16} & 90.64 & -1.52 & 0.547 & 0.171 & 0.31 \\
DermaMNIST & D8-W512 & 75.21 & 75.41 & 0.20 & 89.98 & \textbf{91.26} & 1.28 & 1.799 & 0.573 & 0.32 \\
DermaMNIST & D12-W256 & 75.21 & 75.26 & 0.05 & 89.91 & 88.03 & -1.89 & 0.798 & 0.203 & 0.25 \\
DermaMNIST & D12-W512 & \textbf{76.06} & 74.66 & -1.40 & 90.18 & 81.50 & -8.68 & 2.668 & 0.700 & 0.26 \\
\midrule
OCTMNIST & D5-W256 & \textbf{70.70} & 67.30 & -3.40 & 91.08 & 89.36 & -1.73 & 0.357 & 0.145 & 0.41 \\
OCTMNIST & D5-W512 & 69.90 & \textbf{68.30} & -1.60 & \textbf{91.49} & \textbf{90.13} & -1.36 & 1.145 & 0.476 & 0.42 \\
OCTMNIST & D8-W256 & 70.40 & 66.40 & -4.00 & 90.68 & 86.48 & -4.20 & 0.545 & 0.169 & 0.31 \\
OCTMNIST & D8-W512 & 68.30 & 64.50 & -3.80 & 91.34 & 86.83 & -4.51 & 1.797 & 0.571 & 0.32 \\
OCTMNIST & D12-W256 & 67.70 & 65.30 & -2.40 & 90.90 & 87.63 & -3.27 & 0.796 & 0.201 & 0.25 \\
OCTMNIST & D12-W512 & 68.40 & 65.80 & -2.60 & 89.48 & 84.74 & -4.74 & 2.665 & 0.698 & 0.26 \\
\midrule
PneumoniaMNIST & D5-W256 & 88.30 & \textbf{89.42} & 1.12 & 94.88 & \textbf{93.93} & -0.95 & 0.357 & 0.145 & 0.41 \\
PneumoniaMNIST & D5-W512 & 87.18 & 88.78 & 1.60 & 95.16 & 92.95 & -2.21 & 1.145 & 0.476 & 0.42 \\
PneumoniaMNIST & D8-W256 & 88.46 & 88.14 & -0.32 & \textbf{95.79} & 91.76 & -4.03 & 0.545 & 0.169 & 0.31 \\
PneumoniaMNIST & D8-W512 & 88.30 & 88.94 & 0.64 & 94.99 & 92.44 & -2.55 & 1.797 & 0.571 & 0.32 \\
PneumoniaMNIST & D12-W256 & \textbf{90.06} & 88.94 & -1.12 & 94.99 & 86.84 & -8.15 & 0.796 & 0.201 & 0.25 \\
PneumoniaMNIST & D12-W512 & 87.34 & 88.94 & 1.60 & 94.78 & 87.54 & -7.23 & 2.665 & 0.697 & 0.26 \\
\midrule
RetinaMNIST & D5-W256 & \textbf{58.75} & 55.50 & -3.25 & 73.63 & 71.94 & -1.69 & 0.359 & 0.147 & 0.41 \\
RetinaMNIST & D5-W512 & 56.50 & 56.00 & -0.50 & 74.15 & 73.20 & -0.95 & 1.149 & 0.479 & 0.42 \\
RetinaMNIST & D8-W256 & 56.00 & 56.50 & 0.50 & 73.60 & 74.07 & 0.47 & 0.547 & 0.171 & 0.31 \\
RetinaMNIST & D8-W512 & 56.25 & \textbf{57.00} & 0.75 & \textbf{75.06} & 73.00 & -2.07 & 1.799 & 0.573 & 0.32 \\
RetinaMNIST & D12-W256 & 55.50 & 55.75 & 0.25 & 74.54 & 73.27 & -1.28 & 0.798 & 0.203 & 0.25 \\
RetinaMNIST & D12-W512 & 55.25 & 56.75 & 1.50 & 74.34 & \textbf{74.44} & 0.10 & 2.668 & 0.700 & 0.26 \\
\bottomrule
\end{tabular}%
}
\end{table*}

For reproducibility, we report our hyperparameters used to train the MLP-Mixers in Table~\ref{tab:mlpmixer_setup}. 

One remaining question on MF is how computationally efficient MF is when compared to BP. To assess this, we measured the training wall time per epoch together with the epoch at which the best test accuracy was attained. We summarize the hardware configuration and wall-time comparison for the D5-W256 MLP-Mixer on COIL-100 in Tables~\ref{tab:mlpmixer_walltime_coil100} and~\ref{tab:mlpmixer_walltime_hardware}. All runs used the same set of hyperparameters as documented in Table~\ref{tab:mlpmixer_setup}.

Since MF demonstrates improved speed on CPU, we further assess its CPU efficiency across the other MLP-Mixer configurations. The corresponding results are presented in Table~\ref{tab:mlpmixer_cpu_arch_sweep_coil100}.

These results show that MF has a hardware-dependent efficiency profile. On CPU,
MF consistently reduces the per-epoch wall time across all tested MLP-Mixer
configurations, with MF/BP ratios between 0.887 and 0.931. This suggests that
MF can be computationally attractive in CPU-bound or resource-constrained
settings, where reducing memory usage and avoiding full backpropagation may be
more important than maximizing GPU throughput.

However, the GPU result shows the opposite trend: for the D5-W256 MLP-Mixer on
COIL-100, MF is slower than BP, with an MF/BP time ratio of 1.18. We therefore do
not claim that MF is universally faster than BP. Instead, the results indicate
that MF's efficiency depends on the hardware setting and implementation regime.
BP benefits from highly optimized GPU kernels and mature automatic
differentiation pipelines, whereas MF's layer-wise local objectives introduce
additional classifier-head computations that may not be as efficiently fused or
parallelized on GPU in our current implementation. We leave a detailed investigation of these hardware-dependent efficiency behaviors to future work.

The MLP-Mixer experiments require approximately 10 GPU days to complete, using the same setup described in Appendix~\ref{app:mlp-experiments} and Table~\ref{tab:mlpmixer_walltime_hardware}. All inputs were normalized using dataset-specific channel statistics.
\begin{table}[t]
\centering
\small
\setlength{\tabcolsep}{4pt}
\renewcommand{\arraystretch}{1.08}
\caption{Hyperparameter setup used for the logged MLP-Mixer benchmark runs.}
\label{tab:mlpmixer_setup}
\begin{tabular}{@{}p{0.34\linewidth}p{0.58\linewidth}@{}}
\toprule
Hyperparameter & Value / setting \\
\midrule
Depth & $\{5, 8, 12\}$ \\
Embedding dimension $d$ & $\{256, 512\}$ \\
Patch size & CIFAR-10/CIFAR-100: 4; TinyImageNet: 8; COIL-100: 16; PathMNIST: 7 \\
Dropout & 0.1 \\
Optimizer & AdamW \cite{loshchilov2019decoupledweightdecayregularization} \\
Learning rate & $3\times10^{-4}$ \\
Weight decay & 0.05 \\
Scheduler & Cosine annealing \\
Training epochs & 480 \\
Early stopping & patience $=10$, minimum delta $=10^{-4}$ \\
Batch size & 128 \\
Mixed precision & Enabled \\
\bottomrule
\end{tabular}
\end{table}

\begin{table*}[t]
\centering
\small
\setlength{\tabcolsep}{4.5pt}
\renewcommand{\arraystretch}{1.10}
\caption{Wall-time and accuracy comparison of BP and MF on COIL-100 using a D5-W256 MLP-Mixer. Best test accuracy is reported together with the epoch at which it is attained. Wall-time values are in seconds and are computed as means over the first 50 epochs.}
\vspace{6pt}
\label{tab:mlpmixer_walltime_coil100}
\resizebox{\textwidth}{!}{%
\begin{tabular}{l l l l r r r r r r r}
\toprule
Device  & BP Time & MF Time & MF/BP Time & BP Best Acc & BP Best Epoch & MF Best Acc & MF Best Epoch  \\
\midrule
CPU  & 18.14 & 15.33 & 0.85 & 99.73 & 16 & 99.00 & 35  \\
GPU  & 0.55 & 0.64 & 1.18 & 99.40 & 28 & 99.00 & 35  \\
\bottomrule
\end{tabular}%
}
\end{table*}

\begin{table}[t]
\centering
\small
\setlength{\tabcolsep}{5pt}
\renewcommand{\arraystretch}{1.08}
\caption{Hardware and software environment used for the MLP-Mixer wall-time experiments.}
\label{tab:mlpmixer_walltime_hardware}
\begin{tabular}{@{}ll@{}}
\toprule
Component & Specification \\
\midrule
OS & Ubuntu 22.04 LTS \\
Kernel & 6.8.0-94-generic \\
CPU & 13th Gen Intel Core i9-13900KF \\
CPU threads & 32 logical CPUs \\
System RAM & 62\,GiB \\
GPU & NVIDIA GeForce RTX 4090 \\
GPU memory & 24\,GB \\
Driver & NVIDIA 590.48.01 \\
CUDA & 13.1 \\
\bottomrule
\end{tabular}
\end{table}

\begin{table}[H]
\centering
\small
\setlength{\tabcolsep}{6pt}
\renewcommand{\arraystretch}{1.08}
\caption{CPU wall-time sweep for MLP-Mixer on COIL-100. Values denote the mean training time per epoch after excluding one warmup epoch from a 3-epoch timing run.}
\label{tab:mlpmixer_cpu_arch_sweep_coil100}
\begin{tabular}{c c
                S[table-format=3.2]
                S[table-format=3.2]
                S[table-format=1.3]}
\toprule
{Depth} & {Width} & {BP (s)} & {MF (s)} & {MF/BP} \\
\midrule
5  & 256 & 17.39  & 15.52  & 0.892 \\
5  & 512 & 61.36  & 57.11  & 0.931 \\
8  & 256 & 26.02  & 24.16  & 0.928 \\
8  & 512 & 95.16  & 85.47  & 0.898 \\
12 & 256 & 39.66  & 35.18  & 0.887 \\
12 & 512 & 138.19 & 124.70 & 0.902 \\
\bottomrule
\end{tabular}
\end{table}

\section{Ethical Statements and Broader Impacts}
\label{app:broder-impacts}
This paper presents an algorithmic study of local learning for neural networks. The work uses publicly available benchmark datasets and does not involve human subjects, crowdsourcing, private data, sensitive personal information, deception, surveillance, or deployment in high-risk decision-making settings. To the best of our knowledge, the research conducted in this paper conforms to the NeurIPS Code of Ethics.

The potential positive impact of this work is that Mono-Forward may reduce training memory requirements in some settings, which could improve the accessibility and energy efficiency of neural-network training, especially for researchers and practitioners with limited computational resources. The work may also contribute to the broader study of biologically inspired and local learning mechanisms.

The potential negative impacts are mainly inherited from downstream uses of neural networks. Since Mono-Forward is a general-purpose learning algorithm, it could be applied to harmful or poorly validated applications in the same way as other neural-network training methods.

\paragraph{Existing assets and licenses.}
This work uses publicly available benchmark datasets and standard machine-learning software libraries. We cite the original papers or official sources for the datasets used in our experiments, including MNIST \cite{deng2012mnist, lecun_mnist}, FashionMNIST \cite{xiao2017fashionmnist}, CIFAR-10 \cite{cifar}, CIFAR-100 \cite{CIFAR100}, COIL-100 \cite{nene96columbia}, MedMNIST \cite{Yang_2023}, and TinyImageNet \cite{5206848}. MNIST is distributed under CC BY-SA 3.0, FashionMNIST is distributed under the MIT license; MedMNIST datasets are distributed under CC BY 4.0; COIL-100 is distributed under Apache License 2.0. For datasets whose source pages do not clearly specify a standard license, such as CIFAR-10/100, and TinyImageNet, we use them only as public research benchmarks, follow the terms of the hosting source where applicable, and do not redistribute them. No existing datasets, pretrained models, or third-party codebases are redistributed with this submission.

\clearpage

\end{document}